\documentclass[11pt]{article}
\usepackage[table]{xcolor}

\usepackage{amsmath,amsfonts,bm}

\def\eqref#1{equation~\ref{#1}}
\def\1{\bm{1}}

\def\vtheta{{\bm{\theta}}}

\def\mP{{\bm{P}}}

\DeclareMathAlphabet{\mathsfit}{\encodingdefault}{\sfdefault}{m}{sl}
\SetMathAlphabet{\mathsfit}{bold}{\encodingdefault}{\sfdefault}{bx}{n}

\newcommand{\E}{\mathbb{E}}

\usepackage{fullpage}
\usepackage{natbib}
\setcitestyle{authoryear,round,citesep={;},aysep={,},yysep={;}}
\usepackage{amsmath}
\usepackage{algorithm}
\usepackage{algpseudocode}
\usepackage[pagebackref=true]{hyperref}
\usepackage{url}
\definecolor{ForestGreen}{rgb}{0.13,0.55,0.13}
\usepackage{cleveref} 
\usepackage{booktabs}
\usepackage{listings}
\usepackage{array} % For the 'm' column type
\usepackage{enumitem} % For numbered lists
\usepackage{bbm} % \mathbbm{1}
\usepackage[draft]{graphicx}
\usepackage[toc,page,header]{appendix}
\usepackage{minitoc}
\usepackage[most]{tcolorbox} 
\usepackage{adjustbox}
\usepackage{wrapfig}
\usepackage{nicefrac}
\usepackage{multirow} % add this in your preamble
\usepackage{caption}
\usepackage{minted} % requires -shell-escape
\setminted{fontsize=\footnotesize, breaklines, autogobble, frame=single}

\tcbuselibrary{theorems}
\newtcbtheorem[number within=section]{prompt}{Prompt}%
{enhanced, breakable, colframe=black!50, colback=gray!10, fonttitle=\bfseries}{pr}

\newcommand{\CommentGreen}[1]{\hfill \textcolor{blue}{// #1}}

\newcommand{\Mistral}{\texttt{Mistral-\allowbreak 7B-\allowbreak Instruct-\allowbreak v0.3}}
\newcommand{\Llama}{\texttt{Llama-\allowbreak 3-\allowbreak 8B-\allowbreak Instruct}}
\newcommand{\llamatwo}{\texttt{Llama-\allowbreak 2-\allowbreak 7B}}
\newcommand{\llamathreeone}{\texttt{Llama-\allowbreak 3.1-\allowbreak 8B}}
\newcommand{\skywork}{\texttt{Skywork-\allowbreak Reward-\allowbreak Llama-\allowbreak 3.1-\allowbreak 8B-\allowbreak v0.2}}

\newcommand{\GPTfour}{\texttt{GPT-\allowbreak 4}}
\newcommand{\GPTfouro}{\texttt{GPT-\allowbreak 4o}}
\newcommand{\GPTothree}{\texttt{GPT-\allowbreak o3-\allowbreak mini}}
\newcommand{\Qwenonepointfive}{\texttt{Qwen1.5-\allowbreak 7B-\allowbreak Chat}}
\newcommand{\Qwentwopointfive}{\texttt{Qwen2.5-\allowbreak 7B-\allowbreak Instruct}}
\newcommand{\llmasajudge}{\texttt{LLM-\allowbreak as-\allowbreak a-\allowbreak judge}}

\newcommand{\GSM}{\texttt{GSM8k}}
\newcommand{\Math}{\texttt{Math1-3}}
\newcommand{\Pronto}{\texttt{ProntoQA}}
\newcommand{\gotwentyfour}{\texttt{Game of 24}}

\definecolor{DarkBlue}{HTML}{0B3D91} % tweak if you want it darker/lighter

\DeclareTextFontCommand{\textboldblue}{\color{DarkBlue}\bfseries}
\newcommand{\boldblue}[1]{\textboldblue{#1}}

\definecolor{myblue}{RGB}{0,60,130} % dark blue
\newcommand{\highlight}{\cellcolor{myblue!15}} % lighter shade so text stays legible

\title{Towards Active Synthetic Data Generation \\ for Finetuning Language Models}

\author{
Samuel Kessler\footnotemark[1] \footnotemark[2],
Menglin Xia\footnotemark[2],
Daniel Madrigal Diaz\footnotemark[2],\\
Dongge Han\footnotemark[2],
Helia Heshemi\footnotemark[2],
Saravan Rajmohan\footnotemark[2],
Victor Ruehle\footnotemark[3] \footnotemark[2],
Jordan T. Ash\footnotemark[3] \footnotemark[4]
}

\date{} % Removes the date

\begin{document}

\maketitle

\footnotetext[1]{Corresponding author: \texttt{samuel.kessler@microsoft.com}.}
\footnotetext[3]{Equal advising.}
\footnotetext[2]{Microsoft.}
\footnotetext[4]{Microsoft Research NYC.}

\begin{abstract}
A common and effective means for improving language model capabilities involves finetuning a ``student'' language model's parameters on generations from a more proficient ``teacher'' model. Termed ``synthetic data'', these generations are often produced before any student finetuning, but some work has considered generating new synthetic samples as training progresses. This paper studies and advocates for the latter case, where data are generated in an iterative, closed-loop fashion that is guided by the current state of the student model. For a fixed budget of generated samples, or a budget in terms of compute spent querying a teacher, we show that this curation of finetuning data affords improved student performance over static generation. Further, while there have been several LLM-specific methods proposed that operate in this regime, we find that simple, inexpensive selection criteria from the active learning literature tend to be most performant. We validate these claims across four mathematical and logical reasoning datasets using four different small language models.
\vspace{-0.8cm}
\end{abstract}

\doparttoc % Tell to minitoc to generate a toc for the parts
\faketableofcontents % Run a fake tableofcontents command for the partocs

\part{} % Start the document part
%\parttoc % Insert the document TOC, this will add a table of contents

\section{Introduction}

% Large Language Models (LLMs) have shown remarkable abilities in a wide variety of reasoning and factual knowledge tasks~\citep{achiam2023gpt, bubeck2023sparks, katz2024gpt}. However, due to their size, they are expensive to serve at inference time~\citep{nvidia_llm_inference_cost_2025}. With the advent of agentic systems that interact with the external world, using tools like web search or database queries, LLMs are poised to become even more ubiquitous in science, technology, and society in general. That said, the tremendous inference cost involved in generating sequences from contemporary language models presents a challenge for realizing the full potential of these agents.

Despite the tremendous cost of inference, Large Language Models (LLMs) have risen to prominence as a result of their remarkable abilities across a wide array of reasoning and factual knowledge tasks~\citep{achiam2023gpt, bubeck2023sparks, katz2024gpt}. As agentic systems capable of interacting with the external world emerge, these models are poised to become even more ubiquitous in science, technology, and society, but the tremendous inference cost presents a challenge for realizing the full potential of these agents.

One way to quell the computational expense associated with LLM inference is to use small language models (SLMs). With orders of magnitude fewer parameters, SLMs are faster, cheaper, and easier to finetune for specialised skills like tool use, making them natural specialists using proprietary data or within bespoke agentic systems~\citep{belcak2025small}.

%enables the automation of certain tasks such as creating research reports~\citep{nakano2021webgpt, openai2025deepresearch} or troubleshooting software system failures~\citep{zhang2024flash}
%\jordan{i believe this is called ``tool use''}. 
%LLM use is only set to increase and so present a further challenge in terms of inference costs.

%SLMs can have $3$ orders of magnitude fewer parameters than LLMs, yet still enjoy compelling performance on more narrow, specific tasks. This makes them faster, cheaper, and easier to finetune for specialised skills like tool use or interface alignment. While LLMs are suited for generalist reasoning and orchestration tasks, SLMs are ideal as specialised agents within agentic systems~\citep{belcak2025small}.

Training language models typically involves three stages: pre-training on large, general-purpose corpora, supervised finetuning (SFT), and reinforcement learning from human feedback (RLHF) or from verifiable rewards (RLVR)~\citep{ouyang2022training}. 
SFT, the focus of this work, is critical for
\begin{wrapfigure}{r}{0.55\linewidth}
    %\vspace{-0.3cm}
    \centering
    %\vspace{-0.2cm}
    \includegraphics[draft=false, width=1.0\linewidth]
    {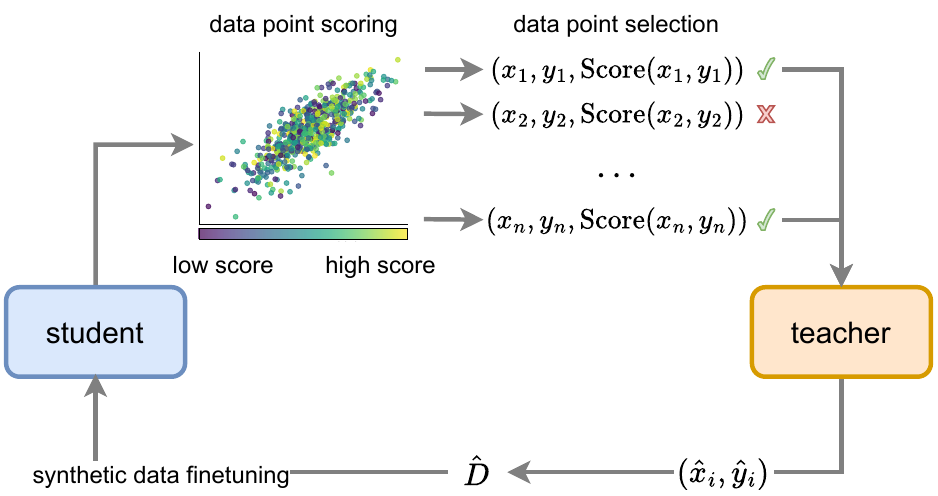}
    \caption{\textbf{Overview of iterative synthetic data generation~(\Cref{alg:iter_syn_data_gen})}. The student model guides synthetic data generation by prioritizing which data are used as an example for the teacher model to generate a new synthetic data point~(\Cref{sec:prompt_syn_gen}). The student finetunes on teacher generated synthetic instruction data.\vspace{-0.0cm}}
    \label{fig:headline}
    %\vspace{-0.3cm}
\end{wrapfigure}
adapting a base model to a target distribution, and is especially common when training SLMs to improve their task-specific performance.

In practice, real-world data for SFT can be hard to obtain, or may lack desirable properties such as chain-of-thought reasoning~\citep{wei2022chain}. Consequently, a typical strategy involves synthesizing a corpus of question and answer pairs from a larger, more capable model~\citep{mitra2024orca, liu2024evolving}. This process usually begins with a small seed dataset, which a teacher LLM uses to produce supplementary synthetic samples before the student SLM is finetuned on the resulting sequences.

Still, evidence suggests that generating a large, static synthetic dataset is frequently wasteful, as it can often be drastically pruned with little to no degradation in trained model capabilities %since we can prune a large proportion of the data and obtain the same capabilities as training on the entire synthetic dataset
~\citep{chen2023alpagasus, zhou2024lima}. As such, this paper explores an iterative, targeted approach to synthetic data generation that is student-aware and improves data efficiency---achieving stronger performance under a fixed data generation budget than naive, static generation---thereby yielding a superior performance–training-set-size Pareto frontier (see~\Cref{sec:prelinaries} for a formal definition).

To facilitate productive learning, this work studies how we can effectively cater to the state of the student model and guide synthetic data generation by a teacher LLM via prompting~\citep{mitra2024orca, liu2024evolving, luo2023wizardmath}. This results in an iterative scheme, where the updated student can be reused to guide further teacher-generated samples~(\Cref{fig:headline}). %data generation. 
%In this way we demonstrate data efficiency benefits.
Prior work has considered this paradigm by prioritizing incorrect student answers~\citep{lee2024llm2llm} and using \llmasajudge{} scoring~\citep{jiang2023lion}, but they do not draw upon the vast active learning and data selection literature. Instead, this paper advocates for the generation of data that are conditioned on samples that have been prioritized by an active learning algorithm.
%We generate data by prompting the teacher model with prioritized data points. At the next iteration the student model will prioritize a new set of data points for downstream synthetic data generation. 
The resulting dataset enables more effective and data efficient finetuning of the SLM student model (see~\Cref{sec:results} for evidence supporting this claim).

Our work makes the following contributions: 
\begin{itemize}
    \item We provide a \boldblue{benchmark study for iterative synthetic data generation rooted in prior work on active learning and data selection, and compare to static synthetic dataset generation}. We show improvements in data efficiency when comparing to generating a single large synthetic instruction dataset, which is a typical approach to student post-training~\citep{mitra2024orca, luo2023wizardmath}.
    \item We compare a range of methods for selecting samples for seeding synthetic data generation, including those that favour uncertainty, diversity, or difficulty. We conclude that \boldblue{simple methods rooted in active learning, such as using the loss of the student's own prediction are most data efficient.} In contrast, expensive and contemporary methods that use an LLM to judge the difficulty and quality of data, i.e. \llmasajudge{}~\citep{zheng2023judging, jiang2023lion}, surprisingly underperform when compared to simpler alternatives.
    \item We show that synthetic data generation is, to a certain extent, steerable: \boldblue{properties of teacher-generated synthetic data resemble those used to seed the generation process.} If the student selects challenging data---measured as samples that induce high student loss---the teacher generates data with correspondingly high loss on average. Given this relationship, and the fact that recent, specialised, LLM-based strategies often fall short, we argue that research in selection strategies is a fruitful and underexplored avenue for advancing the performance of small language models.
\end{itemize}

\section{Preliminaries}
\label{sec:prelinaries}

\paragraph{Notation.} We use $i$ to index a datapoint in a dataset and $j$ to index a token's position in the sequence. In our framework, learning happens iteratively, where synthetic samples are acquired from the teacher, the student trains on the new, larger dataset, and the process repeats. We use $t$ to index the iteration of synthetic data generation. We denote question and answer pairs $z = (x, y)$, from a dataset of size $n$ drawn from a ground truth distribution $P$: $D_0 = \{z\}_{i=1}^{n} \sim P$. We use the terms ``question'' and ``instruction'' interchangeably for $x$, and ``answer'' and ``response'' interchangeably for $y$. The rationales or chain-of-thought steps~\citep{wei2022chain} are incorporated into the answers $y$, however some datasets are comprised of answers without including chain-of-thought steps. A model $f_{\vtheta}(\cdot)$ with parameters $\vtheta$ generates an answer $\hat{y}$ given a question $x$: $\hat{y} = f_{\vtheta}(x)$. Synthetic questions and answers are denoted $\hat{z} = (\hat{x}, \hat{y})$. Text is encoded into tokens, we denote $V$ as the vocabulary and each token is an indicator vector $\{0, 1\}^{|V|}$. SFT involves minimizing the next-token prediction loss, the length-normalized cross-entropy, over answer tokens given a question, $$\mathcal{L}(z, \vtheta) =  \nicefrac{-1}{|y|} \sum_{j = 1}^{|y|} y_j\log f_{\vtheta}(x, y_{<j}).$$ The model $f_{\vtheta}(\cdot)$ autoregressively generates the next token $\hat{y}_{j}  = f_{\vtheta}(x, \hat{y}_{<j})$ in the sequence.

\paragraph{Data Efficiency.} For a fixed number of samples, if better generalization performance can be achieved by training on one subset of a larger dataset than on another, the former can be considered more data efficient. Formally, let $P$ be the true data distribution over our data $z = (x,y)$. For a selection algorithm $\phi$ that produces a dataset $S_n^{\phi} = \left\{ z_{i} \right\}_{i=1}^{n} \overset{\phi}{\sim} P$, model parameters $\vtheta_n^{\phi}$ result from minimizing the loss over $S_n^{\phi}$. We define the performance, accuracy for example, on a single sample as\looseness=-1
\begin{align}
    \text{perf}_{\phi}(z, \vtheta^{\phi}_{n}) =  \bm{1} \left\{y = f_{\vtheta_{n}^{\phi}}(x) \right \},
\end{align}
and the expected performance as
\begin{align}
    \text{perf}_{\phi}(n) = \E_{z \sim P}\E_{S_{n}^{\phi} \sim P} \left[\text{perf}_{\phi}(z, \vtheta_n(S_n^{\phi}))\right].
\end{align}
Assuming a monotonic increase in performance with $n$, for some target performance $\tau$, the sample complexity can be defined as
\begin{align}
    N_{\phi}(\tau) = \inf \left\{n : \text{perf}_{\phi}(n) \geq \tau\right\},
\end{align}
which measures the smallest $n$ such that $\text{perf}_{\phi}(n) \geq \tau$. For a fixed architecture $f(\cdot)$, algorithm $\alpha$ is more data-efficient than algorithm $\beta$ at level $\tau$ only if $N_{\alpha}(\tau) < N_{\beta}(\tau)$ or if, for a fixed $n$,  $\text{perf}_{\alpha}(n) >\text{perf}_{\beta}(n)$.

\section{Related Work}

\paragraph{Distillation.} Fitting models on synthetic datasets composed of pairs $z = (x, \hat{y})$ of sequences, where $\hat{y}$ is produced by a teacher model conditioned on separately available prompts $x$---often referred to as distillation~\citep{hinton2015distilling}---has been shown to be extremely effective in improving capabilities of SLM student models~\citep{alpaca, peng2023instruction, team2024gemma}. \looseness=-1
\vspace{-0.1cm}
\paragraph{Synthetic question and answer generation.} Going one step further, we can generate \textit{both} questions \textit{and} answers: $\hat{z} = (\hat{x}, \hat{y})$. SFT on synthetic question-answer pairs results in improved capabilities without being restricted by small seed dataset sizes~\citep{mitra2024orca}. Much like in the distillation setting, generating a question-answer pair only requires prompting the teacher model with a seed data point~\citep{liu2024evolving, luo2023wizardmath, zeng2024skywork}. \looseness=-1
\vspace{-0.2cm}
\paragraph{Selective question and answer generation.} Synthetic datasets are known to be compressible---synthetic samples filtered by high \llmasajudge{}~\citep{chen2023alpagasus} values or low student loss~\citep{li2023quantity}, for example, obtain the same performance as finetuning on the entire unpruned corpus. To remedy this inefficiency, rather than generating a large static synthetic dataset and then filtering, we can instead carefully select the seed data used to generate the synthetic samples to produce fewer semantically similar sequences. This is effective when distilling on synthetic answers, $(x, \hat{y})$, by balancing correct and incorrect seed data~\citep{liu2024evolving} and conversely by prioritizing high uncertainty seed data~\citep{zhang-etal-2024-elad}. Moreover, data efficiencies have been shown on synthetic question and answer generation by prioritizing incorrect seed data, which is more data efficient than finetuning on the original corpus~\citep{lee2024llm2llm}. \llmasajudge{} selection is also more data efficient than finetuning on public static synthetic datasets~\citep{jiang2023lion, jazbec2024efficient}. We include \llmasajudge{} scoring due to its widespread use and prioritizing incorrectly answered student responses due to its simplicity. It is worth noting that no prior work benchmarks against static synthetic question and answer generation. \looseness=-1
\vspace{-0.1cm}

\subsection{Assigning a Value to Data}
\label{sec:related_work_value}
\paragraph{Active learning.} Our work makes use of ideas from active learning, which seeks to maximise data efficiency by iteratively identifying and prioritising informative samples for labelling~\citep{settles2009active, settles2008analysis}. Classic strategies for active learning include model prediction disagreement~\citep{freund1997selective, houlsby2011bayesian}, uncertainty~\citep{mackay1992information, gal2017deep, kirsch2019batchbald}, and dataset summarization~\citep{sener2017active, mirzasoleiman2020coresets, colemanselection}. Effective, contemporary methods trade-off between predictive uncertainty and sample diversity in a fashion that is commensurate with large neural networks~\citep{ash2021gone, saran2023streaming}. We consider language model-aligned variations of two popular methods for active learning: uncertainty sampling~\citep{settles2008analysis}, and BADGE, a more modern algorithm that trades-off between predictive uncertainty and the diversity of selected data~\citep{ash2019deep}.\looseness=-1

\paragraph{Data selection.} Related methods aim to estimate the value of data to guide selection, typically using a labelled dataset $(x, y)$. Data can be valued using Shapley values~\citep{ghorbani2019data}, influence functions~\cite{koh2017understanding} or by matching training data to evaluation datasets~\cite{just2023lava, kessler2024sava}; these methods have shown limited effectiveness for language modelling. LLMs have been used to score data points~\citep{zheng2023judging} and for selecting question-answer samples for SFT~\citep{liu2023makes, jiang2023lion, chen2023alpagasus}. Still, it has been shown that LLM scores exhibit biases that hinder their effectiveness in this setting~\citep{xiong2023can, dorner2024limits, panickssery2024llm}. Alternative approaches use training loss or gradient norms with respect to student parameters as an estimate of learning progress~\citep{loshchilov2015online, katharopoulos2018not, jiang2019accelerating, li2023quantity, mindermann2022prioritized, evans2024bad, dai2025data}. However, this has shown limited data efficiency for language models~\citep{kaddour2023no}. Reward models are commonly used to select data points for SFT~\citep{cao2023instruction, dubey2024llama}. This work focuses on reward selection because of its popularity. \looseness=-1

\section{Iterative Synthetic Data Generation}
\label{sec:iter_synthetic_data_gen}

The general iterative synthetic data generation process studied in this paper is shown in~\Cref{alg:iter_syn_data_gen}~\citep{jiang2023lion, lee2024llm2llm}. We expand upon the algorithm's design choices in the next sections. %Specific selection algorithms $\phi$ are described below.% can be composed into a scoring and selection functions with the exception of BADGE, described below. 
Most of these methods can be thought of as explicitly scoring each sample with a value $\{s_i\}_{i=1}^{n}$ where $n = |D_0|$ and $D_0$ is the initial question-answer seed dataset. In these cases, we can select $m = |\bar{D}_t|$ points with the highest scores equivalent to selecting the ``hardest'' points, with the highest uncertainty for instance (described in the next section), which is sometimes called ``argmax'' selection $\bar{D}_t = \text{argmax}_{m} \,\left \{s_i\right\}_{i=1}^{n}$. For completeness, we ablate these decisions, for example instead selecting the ``easiest'' points with lowest uncertainty, and sampling proportionally to scores instead of using argmax selection (\Cref{sec:ablations}). Concrete instantiations of selection strategies $\phi$ are outlined below.

\begin{algorithm}[t]
\caption{Iterative synthetic data generation algorithm for question and answer datasets.}
\textbf{Input:} Seed dataset $D_0$, test set $D_{\text{test}}$, train set $\hat{D}_{-1} = \{\,\}$, student $f_{\vtheta}(\cdot)$, selection algorithm $\phi$.
\begin{algorithmic}[1]
    \For{$t = 0, \ldots, T$}
        \State Generate SLM predictions on $D_t$: $\{ z_i = (x_i, \hat{y}_i) \}_{i=1}^n$ where $x_i \in D_0$ and $\hat{y} = f_{\vtheta}(x)$.
        \State Select data subset: $\bar{D}_t = \phi(D_t)$. 
        \Comment{See \Cref{sec:selection_algorithms} for details.}
        \State Generate synthetic dataset: $\hat{D}_t = \text{Generate}(\bar{D}_t)$. \Comment{See \Cref{sec:prompt_syn_gen} for details.}
        \State SFT on $f_{\vtheta}(\cdot)$ using $\hat{D}_t := \hat{D}_t \cup \hat{D}_{t-1}$ and evaluation on $D_{\text{test}}$.
    \EndFor
\end{algorithmic}
\label{alg:iter_syn_data_gen}
\end{algorithm}

\subsection{Selection Algorithms}
\label{sec:selection_algorithms}
\paragraph{Uncertainty sampling.} A common method in the active learning literature is uncertainty sampling, which, for non-sequential classification models, prioritizes data whose probability mass on the most likely class predicted by the model is smallest~\citep{mackay1992information}. In the sequential, Transformer-based setting, we can score a data point with the loss of the response tokens under the student $f_{\vtheta}(\cdot)$ with parameters $\vtheta$ as $\mathcal{L}(z_i, \vtheta)$~\citep{settles2008analysis}. When the targets used to produce a loss are the model's own generations, this score reflects an uncertainty in the produced sequence. Note that our setting gives us access to the ground-truth label associated with $x$ as well, and thus allows us to compute a true loss in a fashion commensurate with conventional model training~\citep{loshchilov2015online}. Interestingly, we find this to be less effective empirically than using the former, uncertainty-based approach~(\Cref{sec:ablations}).

\paragraph{Reward scores.} Using the student's own generated sequence $\hat{y}$, a common method for scoring data is to obtain a prediction from a separate reward model $r(x, \hat{y})$. Resulting scores can be interpreted as the quality of the student's response, and indicative of its competence on questions of this sort in general. We are not limited to using the student's predictions, and can instead obtain a reward for the ground truth answer $y$~\citep{dubey2024llama}. In this manner, rewards capture the difficulty of the data, but this score has no dependence on the student model---we find that using $r(x, y)$ underperforms using $r(x, \hat{y})$ empirically for this reason~(\Cref{sec:ablations}).

%A major drawback of this approach, which will be discussed in more detail later, is that it requires substantial compute in comparison to methods that do not use external models. That is, reasonable comparison between methods, either in terms of the compute budget used or the number of queries sent to a larger model, must account for this additional cost. \jordan{is the only difference between this and the method below that one is an LLM and one is a reward model?} \sam{yes, but also we can get good rms in SLM size} \jordan{also, i don't understand using this on ground truth sequences---does anyone do that? why would it be helpful?} \sam{it is used for Llama3 SFT: 'the reward model is then used to perform rejection sampling on our human annotation prompts.'}

%distinct from the student model, can be used for scoring: $s_i = r(z_i)$. Similarly to loss scoring, we can use the ground truth answer $z = (x, y)$ to give an indication of correctness of the answer and reasoning steps. Alternatively, we can score the student's reasoning steps and final answer $z = (x, \hat{y})$, this reflects the student's ability.

\paragraph{\llmasajudge{} scores.} We can also leverage the reasoning ability of an LLM teacher model to score an SLM's predictions. This strategy asks the LLM teacher to score the detail, quality and correctness of the student's answer and reasoning with a value between $[1, 10]$. In particular, we use pairwise \llmasajudge{} scoring which has been shown to be most effective~\citep{zheng2023judging}. Two separate answers are given for the teacher to decide which it prefers by providing scores for both: $s_i^t, s_i = \texttt{LLM}(\hat{y}^{t}_i, \hat{y}_i, x_i)$ where $\hat{y}^t_i = \texttt{LLM}(x_i)$ is teacher's answer, $s_i^t$ is the score for the teachers answer and $\hat{y}_i$ the student answer. This is expensive, as it requires the teacher to produce an answer in addition to scoring.

\paragraph{BADGE.} Batch Active learning by Diverse Gradient Embeddings (BADGE) is a two-stage active learning algorithm. It first represents all candidate data using the last-layer gradient of the loss induced by treating the generated sequence as ground truth, $\nabla_{ \vtheta_{o}} \mathcal{L}(\hat{y}=f_{\vtheta}(x))$, where $\vtheta_{o}$ are output-head parameters. BADGE then approximately samples from a $k$-DPP to identify gradients that are both high-magnitude and diverse (note that high-magnitude gradients are high-loss generations, suggesting high predictive uncertainty)~\citep{ash2019deep}. Like in uncertainty sampling, our setting allows us to use ground-truth target sequences, which would make these gradient representations of the sort used for optimization, but we found that using generated sequences resulted in better performance. Because the un-embedding layer of a Transformer is typically extremely large, we use a sparse random projection to efficiently reduce dimensionality while preserving geometric  relationships~\citep{johnson1984extensions}.\looseness=-1

%Performs diversity selection with the \texttt{kmeans++} algorithm~\Cref{alg:kmeanspp} which selects points furthest from cluster centres using an $\normltwo$ norm. BADGE uses the gradient of the loss of student's prediction with respect to the final layer of the student. The gradient embeddings capture the importance of the datapoint in a similar manner to the loss and the \texttt{kmeans++} algorithm ensures that we sample a diverse subset of the data. The gradient embeddings can be thought of as scores $s_i := \mG_{\vtheta_o}(x) \in \mathbb{R}^{d \times |V|}$ are calculated as: $\mG_{\vtheta_o}(x) = \nabla_{\vtheta_{o}} \mathcal{L}(z, \theta) = \nabla_{ \vtheta_{o}} \mathcal{L}(\hat{y}=f_{\vtheta}(x))$ where $\vtheta_o$ are the parameters of the final output head, $|V|$ is the size of the vocabulary, $d$ is the student model dimension size. In practice the dimensionality of the gradient embeddings of the final layer is too large for our hardware so we project the embedding dimension down $\vg = \mA^{\top} \text{vec}(\mG_{\vtheta_o}(x)) \in \mathbb{R}^{d}$ using a random projection $\mA \in \mathbb{R}^{(d \cdot |V|)\times d}$~\citep{li2006very}.

\subsection{Prompt-based Synthetic Data Generation}
\label{sec:prompt_syn_gen}
Selected data points $\bar{x}_i \in \bar{D}_t$ are added to a synthetic data generation prompt for the LLM teacher model to generate a synthetic question $\hat{x}_i$~\citep{xu2024wizardlm, mitra2024orca, jiang2023lion, lee2024llm2llm}. The teacher is then prompted to produce chain-of-thought reasoning and a final answer for $\hat{y}_i$. We generate a synthetic data point $\hat{z}_i = (\hat{x}_i, \hat{y}_i)$ using $\hat{x} = \texttt{LLM}(\bar{x}_i)$ and $\hat{y}_i = \texttt{LLM}(\hat{x}_i)$. So $\hat{D}_t = \text{Generate}(\bar{D}_t) = \left\{\hat{x}_i = \texttt{LLM}(\bar{x}_i), \hat{y}_i = \texttt{LLM}(\hat{x}_i)\right\}_{i=1}^{m} \text{, where}\, \bar{x}_i\sim\bar{D}_t$. For details on prompts used for each dataset see~\Cref{sec:prompts}.

\section{Experiments}

This section empirically probes the data efficiency of iterative synthetic data generation against static data generation, and provides recommendations for scoring and selection design choices for data efficiency. \boldblue{We find that prioritizing challenging data, as measured by the student's loss on its own generations, to be at least as data efficient as teacher-based LLM scoring methods, and often more efficient.}\looseness=-1 

LLM-based scoring can behave erratically, particularly for unusual tasks likely outside of the model's training data distribution. Paired with the additional expense of using a large LLM to score data, more general approaches, like uncertainty sampling, appear to be more reliable and effective.\looseness=-1

We further explore this improved data efficiency and \boldblue{show that on average synthetic data inherits some properties from samples used to generate them.} If we select data that are difficult for the student---measured by a high loss or a low reward for example---the resulting synthetic data from the teacher is difficult as well, resulting in lower student accuracies on these generated samples than random selection.\looseness=-1 %Both of these observations explain why selecting data prior to synthetic data generation results in synthetic data that has similar properties to our selected data.

At each iteration $t$ we use a given acquisition algorithm to select $1$k samples $\bar{D}_t$ from $D_t$, before sending each to the teacher model to generate corresponding synthetic data $\hat{D_t}$. These data are appended to synthetic data from all previous iterations before reinitializing the student model and refitting its parameters with gradient descent.

\subsection{Datasets}
\label{sec:datasets}

This section presents results on four distinct reasoning datasets in conjunction with four different models. \GSM{} is a popular mathematics dataset comprised of school level maths problems~\citep{cobbe2021training}, which we use with a \Mistral{} student~\citep{jiang2023mistral}. Similarly, we include the more challenging \Math{} dataset~\citep{hendrycksmath2021}, which is comprised of $5$ distinct levels of question difficulty---we use the easiest levels, $1$ to $3$, to finetune a \Llama{} student~\citep{dubey2024llama}. We further experiment with the logical reasoning dataset \Pronto{}~\citep{saparov2022language}, composed of synthetically generated chain-of-thought style reasoning questions, with a \Qwenonepointfive{} student. 
Finally, we consider the \gotwentyfour{} dataset, which requires finding arithmetic operations to obtain $24$ given $4$ input integers. Here we use a \Qwentwopointfive{} student~\citep{qwen2025qwen25technicalreport}. Specifics are provided in~\Cref{sec:seed_datasets}.

For all datasets except for \gotwentyfour{} we use prompt-based synthetic data generation with a \GPTfouro{} teacher (prompts in ~\Cref{sec:prompts}). Instead, we use backward reasoning: if the answer is $\texttt{13*8-10*8=24}$, for example, we can construct a new question by setting two integers to variables $\texttt{a*b-10*8=24}$ and solving to generate new questions~\citep{jiang2023forward}. We use  a \GPTothree{} teacher for backward reasoning, it qualitatively produces better question-response pairs than \GPTfouro{} (\Cref{sec:prompt_go24}).\looseness=-1

\subsection{Finetuning Setup}
\label{sec:finetuning_setup}

To enable new instruction-following capabilities we finetune our student on synthetic data $\hat{D}_t$, which are appended to synthetic data from all previous iterations $\hat{D}_{<t}$. For efficient training we adapt LoRA layers~\citep{hu2022lora} after each iteration of acquiring data and fitting the model. We avoid warm starting SFT parameters from their pre-trained values and instead use a fresh, random reinitialization~\citep{ash2020warm, springer2025overtrained}. We set the LoRA rank and alpha parameters to the same value (see~\Cref{sec:lora_hparams}) and adapt all linear layers. For optimization we use Adam~\citep{kingma2014adam}, clamp the gradient norm to a maximum of $2.0$, and use a batch size of $24$ with $2$ gradient accumulation steps. The learning rate decays linearly with a warm up period of $15\%$ of the total number of epochs. For \gotwentyfour{} we use a cosine decay learning rate schedule down to a minimum of $1\text{e-}9$~\citep{ni2025offlinelearningforgettingreasoning}. During optimization we perform checkpointing and load the checkpoint with the best performance on a held-out validation set after SFT. We search over learning rates, LoRA ranks and the number of training epochs on this held-out validation set as well~(\Cref{sec:further_experiment_setup}). We use a single 80Gb A100 or H100 GPU for all experiments.

\begin{figure}
    \centering
    \vspace{-0.8cm}
    \includegraphics[draft=false, width=1.0\linewidth]{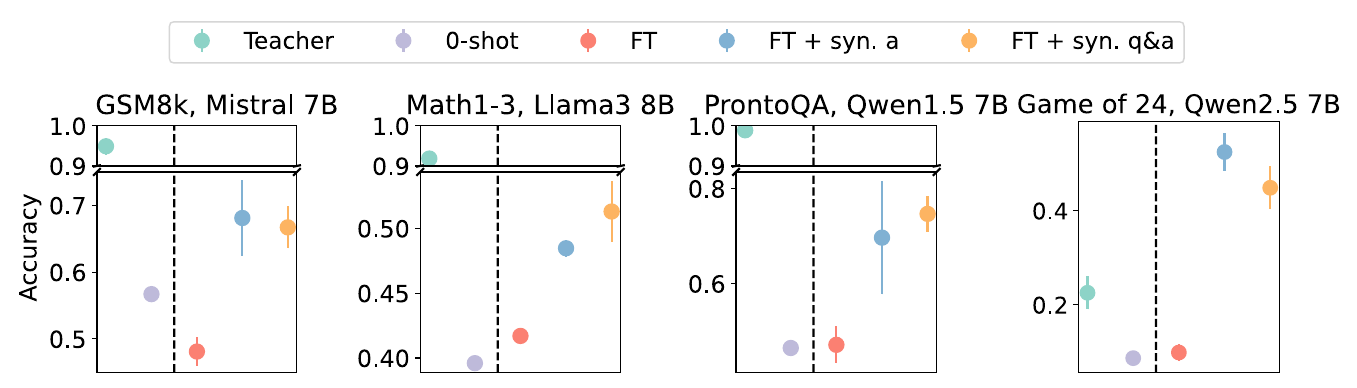}
    \caption{\textbf{SFT performance on $1$k data points for various datasets and SLMs.} We compare the effect of synthetic answer generation and synthetic question and answer generation to using the seed dataset, $D_0$ for SFT. $0$-shot SLM and teacher performances are included for reference. All datasets use a \GPTfouro{} teacher, for \gotwentyfour{} we use a \GPTothree{} teacher. Using synthetic data, either as answers paired with real questions (syn. a) or both questions and answers (syn. q\&a) improve performance past using the seed dataset alone (FT).\looseness=-1}
    \label{fig:synthetic_data_performance}
    \vspace{-0.5cm}
\end{figure}

\subsection{Algorithms}
\label{sec:experimental_setup}

This paper considers a variety of selection algorithms. Prior work has shown that prioritizing ``hard'' samples accelerates learning (\Cref{sec:related_work_value}), which we also find to be the case for iterative synthetic data generation (\Cref{sec:ablations}). This approach prioritizes high-uncertainty data, measured as the model's loss on greedily decoded student generations, which we denote as ``loss (high)'' throughout this section. We also consider a low-reward selection algorithm (``rwd (low)''), also using the student's own predictions, which scores generations using an external model. We use a \skywork{} reward model which is the highest scoring $8$b model on RewardBench~\citep{lambert2024rewardbench} at the time of writing.

We use Lion~\citep{jiang2023lion} as a baseline, which compares the student and teacher answer \llmasajudge{} scores to classify each data point as either an easy or a hard before sampling equally from both sets. For completeness, we also consider a baseline that only samples from the hard set, denoted as \llmasajudge{} (hard)~\citep{jazbec2024efficient}. We use the same prompts for \llmasajudge{} scoring as~\citet{jiang2023lion}.\looseness=-1

We further consider prioritizing data with incorrect student answers, $s_i = \1\{\hat{y}_i \neq y\}$ as a proxy for prioritizing hard samples~\citep{lee2024llm2llm}. In a similar spirit to Lion, we can instead sample evenly from correct and incorrect pools to maintain diversity in the seed data~\citep{liu2024evolving}. Since correct and incorrect scoring requires a verifier and ground-truth answers, we do not compare them to other scoring methods that do not use label information and instead place these supplementary results in~\Cref{sec:incorrect}.

\begin{figure}
    %\centering
    \vspace{-0.5cm}
    \hspace{-0.5cm}
    \includegraphics[draft=false, width=1.05\linewidth]{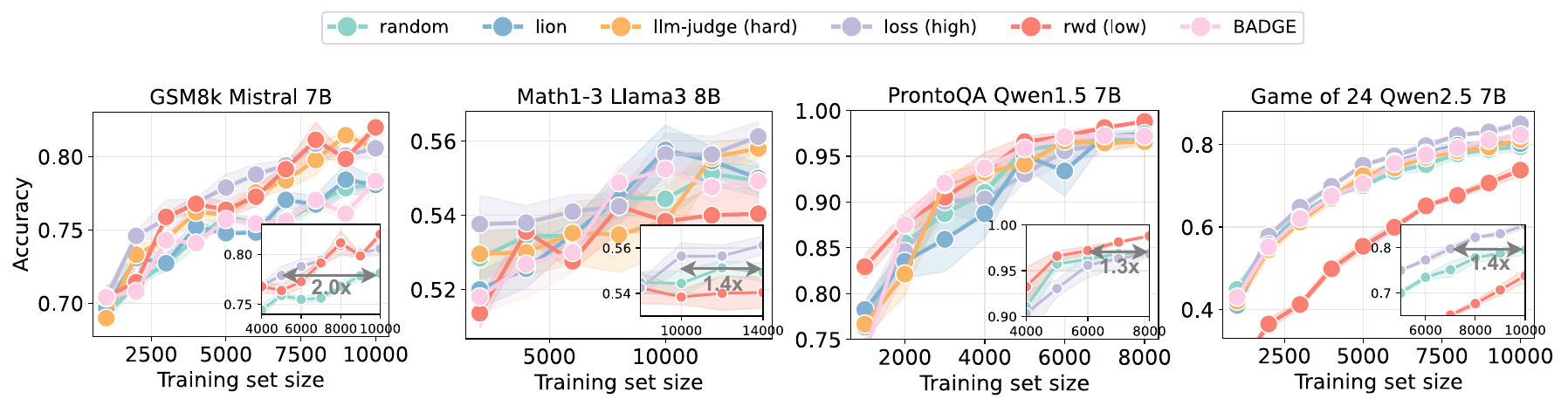}
    \caption{\textbf{Student performance over successive synthetic data iterations with growing training sets}. In all cases, selection based on uncertainty (loss) performs approximately as well as LLM-based scoring strategies (rwd and llm-judge), without requiring additional queries to an LLM. Further, for tasks that are out-of-distribution for the scoring model, like \gotwentyfour, these mechanisms can perform even worse than random sampling.
    Horizontal lines in each inset plot denotes the proportion of data random sampling would require to achieve the same performance as the best active selection strategy in the corresponding experiment.}
    \label{fig:learning_curves}
    \vspace{-0.4cm}
\end{figure}

\subsection{Results}
\label{sec:results}

This subsection presents our main results, which includes (1) performance comparisons between fitting the SLM on seed data and synthetic data (\Cref{sec:syn_vs_seed}), and (2) between standard active learning strategies and more modern, LLM-based alternatives (\Cref{sec:iterative_syn_gen_results}). Further, we (3) analyse synthetically generated data to demonstrate it retains important properties of the original seed data, providing the underlying property of this mechanism affording the effectiveness of active methods (\Cref{sec:syn_data_corr}). 
Throughout this section, note that static generation is equivalent to random sampling of prompts for synthetic data generation in our setting, as it is not conditioned on the current state of the student. Unless stated otherwise, results show the mean and standard deviation over $3$ independent runs\footnote{Website and code: \texttt{https://iterative-sd.github.io/}}.\looseness=-1

\subsubsection{Synthetic Data Improves Performance}
\label{sec:syn_vs_seed}

\boldblue{SFT on synthetic data results in significantly improved capabilities when compared to using the original seed dataset.} ~\Cref{fig:synthetic_data_performance} compares SFT performance on the seed data to synthetic data of equal size, showing a dramatic increase in performance across all datasets when doing SFT on synthetic question-answers pairs. In the same figure, we see large increases in performance when using synthetic answers $z_i = (x_i, \hat{y}_i)$ instead of seed answers $y$, likely due to better formatting and high quality chain-of-thought in synthetic answers. In \gotwentyfour{} there is a small drop in 
\begin{wrapfigure}{r}{0.5\linewidth}
    \vspace{-0.0cm}
    \hspace{-0.3cm}
    \centering
    \includegraphics[draft=false, width=\linewidth]{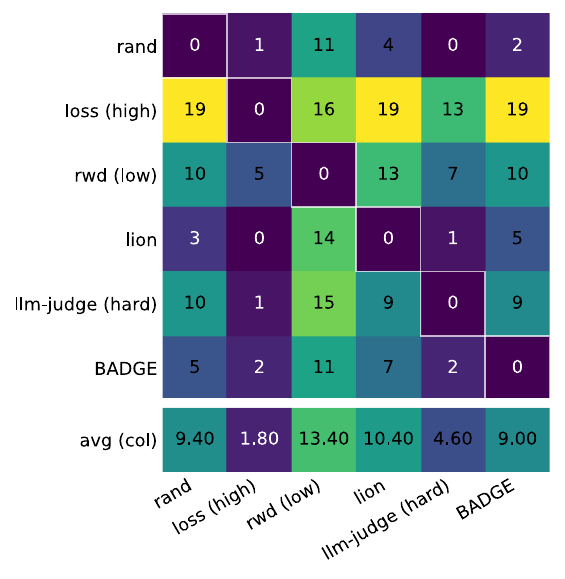}
    \vspace{-0.52cm}
    \caption{\textbf{Pairwise winrate over all datasets and methods.} $\mP_{ij}$ corresponds to the number of times algorithm $i$ outperforms $j$. Overall performance is shown in the last row (lower is better).}
    \label{fig:win_rates}
    \vspace{-1.1cm}
\end{wrapfigure}
performance when training on synthetic questions and answers compared to synthetic answers only, showing that the generation of novel questions by the teacher yields some lower quality synthetic questions. Regardless, next we show how this enables us to scale dataset sizes efficiently.%\looseness=-1

\subsubsection{Iterative Generation is More Data Efficient than Static Generation}
\label{sec:iterative_syn_gen_results}

\boldblue{Active selection is more data efficient than random sampling for generating productive synthetic data, resulting in better performance using fewer samples}. \Cref{fig:learning_curves} shows learning curves, with each plot measuring the test accuracy of a given selection method as a function of the labelling budget; each point is an active learning iteration. As mentioned earlier, random selection, because it is not conditioned on the current state of the student, is equivalent to the typical approach of static generation at the indicated data size. We find that this technique is often outperformed by a student-in-the-loop alternative---horizontal lines indicate the number of additional samples that would be required by static, random sampling in order to achieve the same performance as the best active learner in the corresponding plot (between $1.3\times$ and $2\times$). We find that uncertainty sampling performs roughly as well, and often better than LLM-based scoring methods. For some datasets, like \gotwentyfour{}, reward-based methods do quite poorly, likely as a result of the task being out of distribution for the reward model.

\Cref{fig:win_rates} aggregates performance differences between all selection strategies and model-dataset pairs considered in this paper. Each experiment composing this figure produced a learning curve in \Cref{fig:learning_curves}, with each method producing a different SLM test accuracy for a variety of generation budgets. Here, we aggregate results by measuring which algorithms outperform their peers at each generation budget across all models and datasets.

Specifically, we aggregate results as a pairwise winrate matrix $\mP$. We increment $\mP_{ij}$ if $\1\{\hat{\mu}_i - \alpha \cdot \hat{\text{se}}_{i} > \hat{\mu}_j + \alpha \cdot \hat{\text{se}}_{j} \}$, where $\hat{\mu}_i$ is the sample mean and $\hat{\text{se}}_i$ is the standard error of the performance of algorithm $i$ for a dataset, for a particular dataset size, and $\alpha$ is the confidence level which we set to $1$ (making it a $68\%$ confidence interval). By summing the ``wins'' across the rows and normalizing we can understand how often algorithms are outperformed on average. Column-wise averages are shown in the last row, where lower is better, to understand which algorithm is more data efficient in total. We find that random sampling---equivalent to static generation of data---is outperformed by various other methods that use the student model to guide synthetic data generation~(\Cref{fig:win_rates}).

We can glean from~\Cref{fig:win_rates} that the highest-performing approach is simply uncertainty sampling, using the SLM's loss on its own generations. \llmasajudge{} also tends to be somewhat effective, though by a reduced margin. Interestingly, BADGE and Lion, which both aim to select diverse data, do not perform much better than random sampling~(\Cref{fig:win_rates}). This is likely because synthetic data generation is noisy (\Cref{sec:syn_data_corr}), which diverse selection may exacerbate.

Because of the need to access a teacher model for scoring, \llmasajudge{} is computationally demanding. Assuming that the cost of evaluating the teacher model dominates the cost of evaluating the student, a common assumption in the active learning literature and a reasonable assumption as the number of parameters of the teacher model can be $3$ orders of magnitude larger than the student models we consider. Then, if we consider the total number of teacher input and output tokens as a budget instead of the number of generated samples, Lion and \llmasajudge{} (hard) are 
\begin{wrapfigure}{r}{0.40\linewidth}
    \vspace{-0.4cm} % tweak as needed
    \centering
    \includegraphics[draft=false, width=\linewidth]{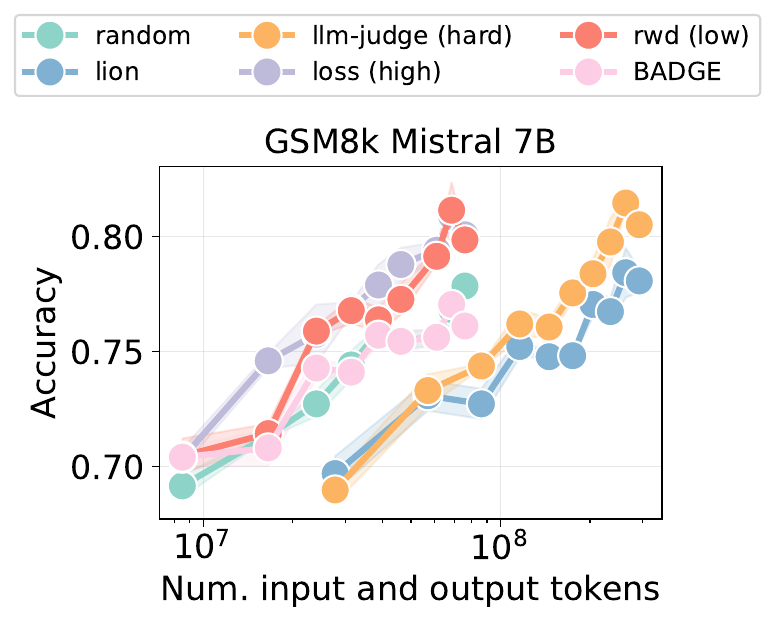}
    \caption{\textbf{Active learning curves on \GSM{}: student performance against the number of teacher input and output tokens}. The total number of input and output tokens are a proxy for the amount of compute used by the teacher for various methods.\looseness=-1}
    \label{fig:learning_curves_num_inputs_tokens}
    \vspace{-0.6cm}
\end{wrapfigure}
far more expensive than other methods~(\Cref{fig:learning_curves_num_inputs_tokens}). \textboldblue{Our results suggest this additional compute is better allocated towards simply generating more synthetic data with a cheaper and more effective selection strategy, like uncertainty sampling.}

\begin{figure}
    \vspace{-0.3cm}
    \centering
    \includegraphics[draft=false,width=1.0\linewidth]{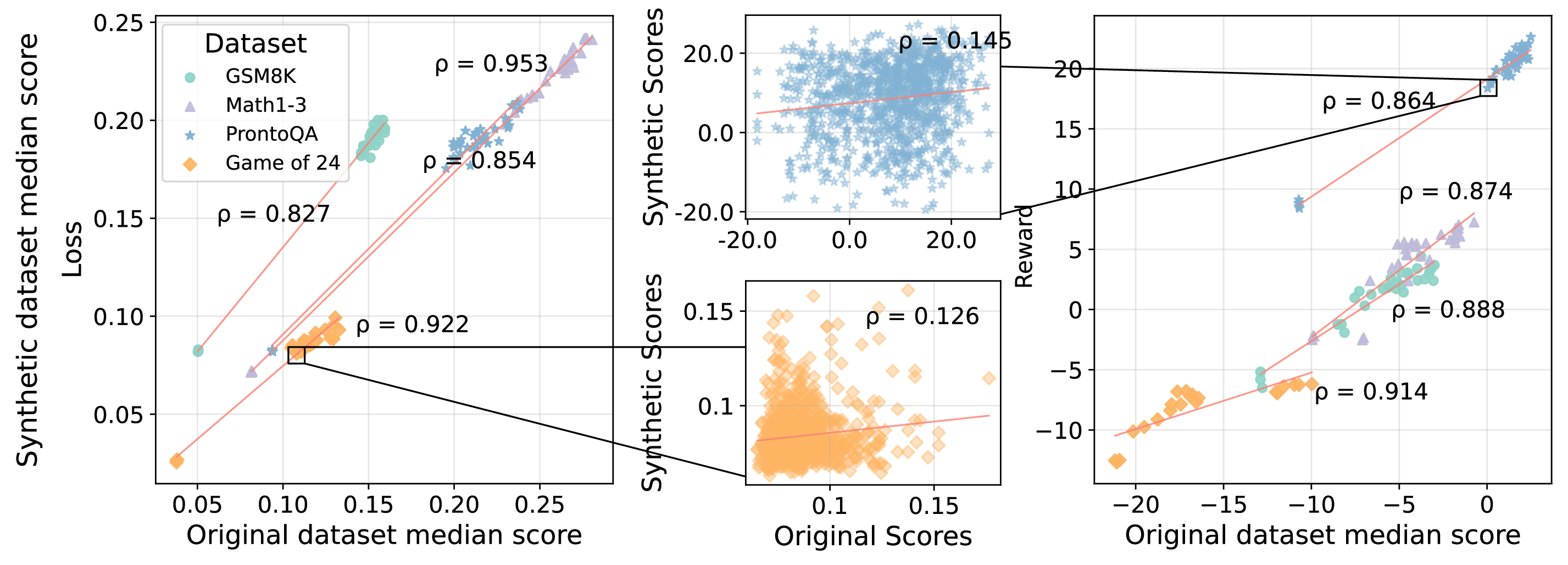}
    \caption{\textbf{The rank correlations between original and synthetic dataset scores from iterative synthetic data generation}. We plot student loss and reward scores and show Spearman's rank correlations ($\rho$) between dataset medians before and after synthetic data generation. We zoom in on relationships at an individual data-point level where there is low correlation between the original and synthetic data point scores (centre). The red line is the line of best fit to the data. All rank correlations are highly significant ($p < 0.001$).}
    \label{fig:synthetic_data_corr}
    \vspace{-0.4cm}
\end{figure}

Reward scoring also requires an external model, but because we can use a reward model that has the same number of parameters as our student, calls to the reward model are generally less expensive than to a teacher---we opt to not treat them in the same way and do not count the number of input tokens to the reward model in~\Cref{fig:learning_curves_num_inputs_tokens}. Overall random selection requires between $33\%$ to $100\%$ more SFT data to obtain the same performance as the best selection methods across all datasets (\Cref{fig:learning_curves}). For $2/4$ of these datasets, iterative synthetic data generation using the loss on the student's own predictions leads to more data efficient results compared to prior works that perform SFT using similarly sized datasets (see \Cref{sec:performance_prior_work} for comparisons).

\subsubsection{Fidelity of Synthetic Data to its Original Data}
\label{sec:syn_data_corr}
Synthetic data generation is a noisy process, perturbing data by rephrasing, complicating or simplifying and adding chain-of-thought rationales. As such the score---either uncertainty (measured by the loss) or data quality (measured by the reward)---of an individual seed sample and its corresponding generated sample appear to have little to do with each other. In aggregate, however, we find high rank correlation between the median score of the seed and generated datasets (\Cref{fig:synthetic_data_corr}). This relationship is the underlying principal governing why careful selection of the seed question-response pairs is important: \textboldblue{Generated samples inherit underlying properties from the data used to produce them}. These attributes, such as the SLM loss, shape the student's optimization trajectory and generalization capabilities.\looseness=-1 

\begin{figure}
    \vspace{-0.8cm}
    \centering
    \includegraphics[draft=false, width=1.0\linewidth]{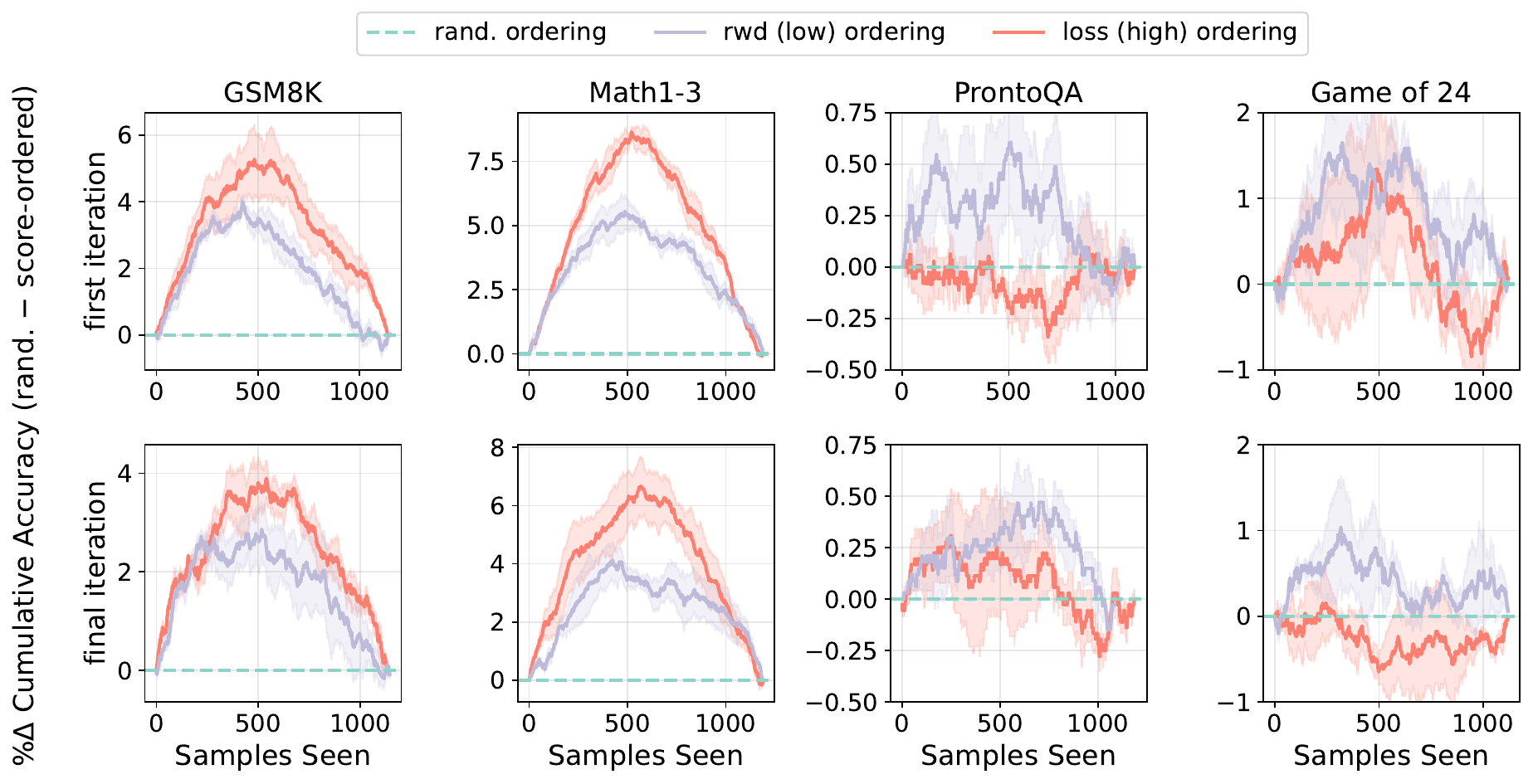}
    \caption{\textbf{The percentage difference in synthetic data cumulative accuracy between samples ordered by score and randomly shuffled.}
    Data are sorted either by uncertainty (high to low) or reward (low to high). Positive values suggest that score ordering picks more difficult synthetic samples  in turn yielding lower accuracies. For each original data point we score it using the student model from the first and final iteration of iterative synthetic data generation (rows). See~\Cref{fig:cum_acc_individual_seeds} for the cumulative accuracies for individual replicates.\looseness=-1}
    \label{fig:cum_acc_difference_aggregated}
    \vspace{-0.5cm}
\end{figure}

Unsurprisingly, samples with high uncertainty, again measured using the student's loss on its own generations, tend to also be samples for which the student model struggles to obtain a correct answer. \Cref{fig:cum_acc_difference_aggregated} sorts samples by this uncertainty and compares the model's accuracy on these data in comparison to a random shuffle, this is equivalent to random selection. Specifically, we plot the cumulative percent difference between the SLM accuracy on samples sorted by loss and the SLM accuracy on randomly ordered data. The curve in \Cref{fig:cum_acc_difference_aggregated} presents this as a function of the number of samples being used in the calculation, and is repeated for scores from a reward model. In both cases, prioritizing data in this fashion often effectively prioritizes low-accuracy samples, as indicated by the curve's positive values. The trend is least clear for the \Pronto{} dataset, which shows reward scoring as positive and uncertainty scoring as neutral or negative---unlike for other datasets, reward scoring was indeed a slightly more effective selection strategy (\cref{fig:learning_curves}). Low reward selection also selects lower-accuracy samples compared to random for the \gotwentyfour{} dataset (\Cref{fig:cum_acc_difference_aggregated}), but still performs more poorly in terms of SFT performance than random~(\Cref{fig:learning_curves}). This is because the reward scorer introduces biases; specifically, it prefers longer responses, an effect that has been observed in other work~\citep{shen2023loose, bu2025beyond}.\looseness=-1

\section{Conclusion and Discussion}
This work shows that student-in-the-loop synthetic data generation yields more data-efficient SLM improvements than static one-shot generation, and that simple, active-learning–inspired criteria for selecting seed examples outperform more elaborate LLM-based judging. We demonstrate that synthetic data are partially steerable, with teacher outputs reflecting the properties of selected seeds. These results highlight the value---and underexploration---of principled, effective selection strategies for advancing SLM training. Limitations are overviewed in~\Cref{sec:limitations}.

\begin{comment}
\section{Reproducibility Statement}

Reproducibility goes to the heart of our study of different selection algorithms for data efficient synthetic data generation. In our study, all the results are stated as means and standard errors over $3$ independent replicates. This has been done in an effort to encapsulate the variance arising from the datasets we use and our experimental setup, and ensures that the performance differences arise due to the choice of selection methods rather than random variation. As a result we use uncertainties to weight our claims resulting a more reproducible study.

We will release source code upon publication.
\end{comment}

\section{Acknowledgements}
We thank Guoqing Zheng for insightful discussions.

\bibliographystyle{plainnat}
\bibliography{main}

\newpage
\appendix
\addcontentsline{toc}{section}{Appendix} % Add the appendix text to the document TOC

\part{Appendix} % Start the appendix part
\parttoc % Insert the appendix TOC

\begin{comment}
\section{The Use of Large Language Models}

We used a coding assistant to help implement and debug our experiments. Regarding paper writing we used LLMs for finding related work, to assist with grammar queries, and to assist in generating the figures in the paper.
\end{comment}

\section{Limitations}
\label{sec:limitations}
\vspace{-.2cm}
\paragraph{Iterative synthetic data generation for finetuning.} We only consider SFT, we do not consider efficient synthetic data generation to accelerate training for RLHF, continual pre-training~\citep{yang2024synthetic} or pre-training~\citep{maini2025beyondweb}, for instance. These are promising directions of future work.

\paragraph{The limits of the teacher.} We assume that the teacher is able to generate high quality questions and answers. For \GSM{}, \Math{} and \Pronto{} the teacher performance is high and so we assume $\hat{z}_i$ is correct. For \gotwentyfour{} we rely on backward reasoning (specific to arithmetic) and a verifier to assess the teacher's synthetic data. We have yet to test the limits of prompt-based synthetic data generation in settings where teacher capabilities fall short.

\paragraph{Data generation is noisy.} We can obtain improved student capabilities using iterative synthetic data generation. However, synthetic data generation is a noisy process where we show that properties of the selected datasets are preserved~(\Cref{sec:syn_data_corr}). However, it is not clear how we can guarantee that synthetic data retains desirable properties from the seed dataset. For example, reward scoring performs poorly for the \gotwentyfour{} since it is biased by long student answers despite also selecting low quality student responses for synthetic data generation as required.
\begin{table}[ht]
    \centering
    \begin{adjustbox}{max width=\linewidth}
    \begin{tabular}{l l c c c}
    \toprule
    \textbf{Model} & \textbf{Dataset} & \textbf{LoRA Rank} & \textbf{Learning Rate} & \textbf{Epochs}\\
    \midrule
    \Mistral & \GSM \, seed &  32 & 1\text{e-}4 & 10 \\
    \Llama & \Math \, seed  & 32 & 1\text{e-}6 & 13 \\
    \Qwenonepointfive{} & \Pronto \, seed & 32 & 1\text{e-}5 & 13\\
    \Qwentwopointfive{} & \gotwentyfour \, seed & 16 & 1\text{e-}5 & 13 \\
    \Mistral & \GSM \, synthetic &  32 & 1\text{e-}4 & 10 \\
    \Llama & \Math \, synthetic  & 64 & 1\text{e-}4 & 13 \\
    \Qwenonepointfive{} & \Pronto \, synthetic & 32 & 1\text{e-}5 & 13\\
    \Qwentwopointfive{} & \gotwentyfour \, synthetic & 16 & 5\text{e-}4 & 30 \\
    \bottomrule
  \end{tabular}
  \end{adjustbox}
  \caption{Optimal hyper-parameters for LoRA fine-tuning for all seed and synthetic datasets.}
  %\vspace{1.5cm}
  \label{tab:optimal_hparams}
\end{table} We have presented an initial study of the ``steerability'' of synthetic data generation. However the ability to add further desirable properties is left for future work.

\section{Additional Experimental Setup Details}
\label{sec:further_experiment_setup}

We introduce additional details of our experimental setup from~\Cref{sec:experimental_setup}. We outline the hyperparameter grid search for SFT.

\subsection{LoRA Hyper-parameter Tuning Setup}
\label{sec:lora_hparams}

To obtain the best hyperparameters for our seed datasets $D_0$ and our synthetic datasets $\hat{D}_t$,
we sweep through a grid of learning rates, number of training epochs and LoRA rank hyper-parameters using $1$k question-answer pairs from the original seed dataset and $1$k question-answer pairs synthetically generated by the teacher model. Refer to~\Cref{tab:optimal_hparams} for the optimal hyperparameters found in our sweep.

\section{Additional Results}

We introduce additional results that support the claims in our main paper. In~\Cref{sec:incorrect}, we introduce the results of prioritizing synthetic data generation using incorrect student predictions~\citep{lee2024llm2llm} and an even number of correct and incorrect student data~\citep{liu2024evolving}. We do not include these results in the main paper for comparison since they require the ground truth answer $y$ for scoring unlike the other active scoring methods considered~(\Cref{sec:iterative_syn_gen_results}). In~\Cref{sec:performance_prior_work}, we compare iterative synthetic data generation with comparable SFT methods from the literature. In~\Cref{sec:syn_data_properties}, we analyse the workings of synthetic data generation to show that despite introducing noise, the synthetic data retains the scores of the original selected seed data in aggregate. Furthermore, in~\Cref{sec:syn_acc_cumsum}, we study how the synthetic datasets which are prioritized by low reward and high loss selection algorithms result in more difficult synthetic datasets since we observe lower student accuracies. In~\Cref{sec:tvd}, we show how the different scorers produce synthetic datasets with different token frequency distributions. These observations explain why selecting data prior to synthetic data generation results in data that has similar properties to our selected data and therefore enhanced student performance upon finetuning. Finally, in~\Cref{sec:ablations}, we compare various design choices for iterative synthetic data generation~(\Cref{alg:iter_syn_data_gen}).

\begin{figure}
    \centering
    \includegraphics[draft=false, width=1.0\linewidth]{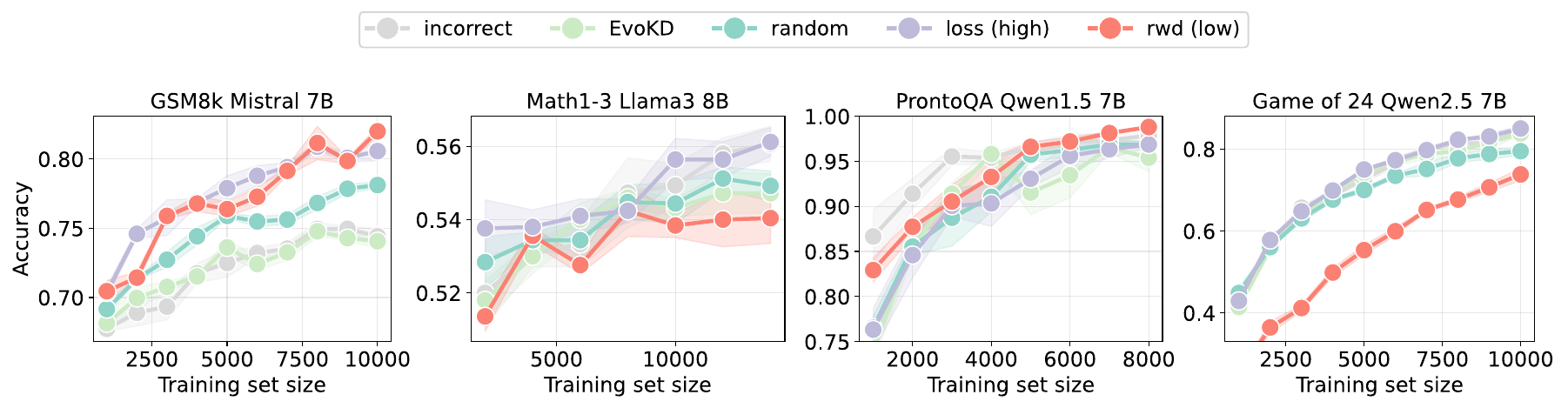}
    \caption{\textbf{Iterative synthetic data generation learning curves, showing student SFT performance after training on synthetic data of increasing size with incorrect~\citep{lee2024llm2llm} and EvoKD~\citep{liu2024evolving} data prioritization}. Each consecutive increase in dataset size corresponds to an iteration of iterative synthetic data generation~(\Cref{alg:iter_syn_data_gen}). Learning curves are across various dataset student model pairs. Curves are an average and standard error over $3$ replicates.}
    \label{fig:learning_curves_incorrect}
    \vspace{-0.4cm}
\end{figure}

\subsection{Prioritizing Incorrect Samples}
\label{sec:incorrect}

\begin{wrapfigure}{r}{0.5\linewidth}
    \vspace{-1.7cm}
    \centering
    \includegraphics[draft=false, width=\linewidth]{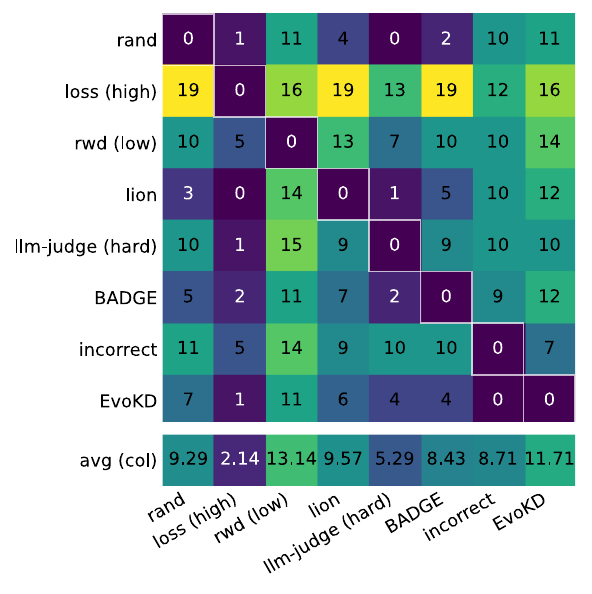}
    \vspace{-0.6cm}
    \caption{\textbf{The pairwise win rate matrix over all datasets and all methods including incorrect prioritization}. Element $\mP_{ij}$ corresponds roughly to the number of times algorithm $i$ outperforms algorithm $j$ including results of incorrect student answer prioritization~\citep{lee2024llm2llm} and EvoKD~\citep{liu2024evolving}. Column-wise averages at the bottom display overall performance (lower is better).}
    \label{fig:win_rates_incorrect}
    \vspace{-0.6cm}
\end{wrapfigure}

\boldblue{Prioritizing incorrect student predictions yields strong performance on all but one of the datasets we consider.} A simple data point scoring mechanism is to assign a $\{0, 1\}$ score for an incorrect or correct answer from the student model. This scoring mechanism requires a verifier or the ground truth answer $y$ 
and so is not directly comparable to the active scoring methods we consider that do not require the ground truth answer for scoring~(\Cref{sec:experimental_setup}). Regardless, we show the results of performing iterative synthetic data generation by prioritizing incorrect samples in~\Cref{fig:learning_curves_incorrect}. For \GSM{} this method severely underperforms other prioritization methods and random sampling.
\begin{table}[h!]
\centering
\vspace{-0.2cm}
\begin{adjustbox}{max width=\linewidth}
\begin{tabular}{lllcc}
\toprule
\textbf{Dataset} & \textbf{Method} & \textbf{LLM} & \textbf{SFT Dataset Size} & \textbf{Performance} \\
\midrule
\multirow{4}{*}{\GSM{}}
    & Teacher & \GPTfouro & n/a & $94.9 \pm 1.1$ \\
    %& Pre-memorization Accuracy~\citep{kang2024learning} & \Llama{} & $10$k & $58^{\dagger}$ \\
    %& Multiagent Debate~\citep{subramaniam2025multiagent} & \Mistral{} & - & $58.4 \pm 2.1$ \\
    & Orca-Math~\citep{mitra2024orca} & \Mistral{} & $10$k & $70.2$ \\
    %& MetaMath~\citep{metamath} & \llamatwo{} & $240$k & $66.5$ \\
    %& MetaMath~\citep{metamath} & \Mistral{} & $240$k & $77.7$ \\
    %& MuMath~\citep{mumath} & \llamatwo{} &  $384$k & $76.2$\\
    %& Xwin-Math~\citep{xwinmath} & \llamatwo{} & $960$k & $82.6$ \\
    %& Xwin-Math~\citep{xwinmath} & \Mistral{} & $960$k & $89.2$ \\
    & OpenMathInstruct~\citep{openmathinstruct} & \Mistral{} & $1.8$M & $80.2$\\
    %& OpenMathInstruct + Self-Consistency~\citep{openmathinstruct} & \Mistral{} & $1.8$M & $86.9$\\
    & \highlight Iterative Synthetic Data Generation (ours) & \highlight \Mistral{} & \highlight $10$k & \highlight $80.6 \pm 1.2$ \\
\midrule
\multirow{2}{*}{\Math{}} 
    & Teacher & \GPTfouro & n/a & $91.8 \pm 0.7$ \\
    %& Pre-memorization Accuracy~\citep{kang2024learning} & \texttt{DeepSeekMath 7B} & $10$k & $47^{\dagger}$ \\
    %& MetaMath~\citep{sun2024easy} & \texttt{Llemma-\allowbreak 7B} & $155$k & 44.1 \\
    %& Multiagent Debate~\citep{subramaniam2025multiagent} & \Llama{} & - & $57.4 \pm 2.2$ \\
    & \highlight Iterative Synthetic Data Generation (ours) & \highlight \Llama{} & \highlight $10$k & \highlight $56.1 \pm 0.9$ \\
\midrule
\multirow{2}{*}{\Pronto{}}
    & Teacher & \GPTfouro & n/a & $98.9 \pm 0.4$ \\
    %& Chain of Thought + Self-Consistency~\citep{li2025policy} & \llamathreeone & n/a & $74.6$ \\
    %& Monte Carlo Tree Search + Self-Consistency~\citep{li2025policy} & \llamathreeone & n/a & $74.2$ \\
    %& Policy Guided Tree Search + Self-Consistency~\citep{li2025policy} & \llamathreeone & n/a & $77.4$ \\
    %& SFT-NL~\citep{zhou2025dissecting} & \Qwentwopointfive{} &  $3.2$k & 97.4 \\
    & \highlight Iterative Synthetic Data Generation (ours) & \highlight \Qwenonepointfive{} & \highlight $8$k & \highlight $96.9 \pm 0.8$
 \\
\midrule
\multirow{3}{*}{\gotwentyfour{}}
    & Teacher & \GPTothree & n/a & $22.6 \pm 1.8$ \\
    %& Chain of Thought + Self-Consistency~\citep{yao2023tree} & \GPTfour & n/a & $9.0$ \\
    %& Tree of Thoughts~\citep{yao2023tree} & \GPTfour & n/a & $74.0$ \\
    & UFT~\citep{ni2025offlinelearningforgettingreasoning} & \Qwentwopointfive{} & $13.7$k & $30.2\pm2.1$ \\
    & \highlight Iterative Synthetic Data Generation (ours) & \highlight \Qwentwopointfive{} & \highlight $6$k & \highlight $85.0 \pm 1.3$ \\
\bottomrule
\end{tabular}
\end{adjustbox}
\caption{\textbf{Iterative synthetic data generation performs comparably to state-of-the-art SFT methods on certain datasets}. The results of iterative synthetic data generation using high loss selection, as this selection method performs the best overall. We compare only to methods that use the same LLM and omit work that relies on larger datasets to achieve higher performance, as we cannot determine whether such gains stem from better techniques or simply from increased data. All SFT methods report the amount of data used for SFT. We report a mean and standard error over multiple seeds for our work, however some baselines only report a single seed.}
\label{tab:prior_works_comparison}
\vspace{-0.0cm}
\end{table}
For \Math{} and \gotwentyfour{} incorrect student answer prioritization is as data efficient as high loss scoring which is the most data efficient method identified in~\Cref{sec:iterative_syn_gen_results}. For the \Pronto{} dataset incorrect answer prioritization obtains results on par with the best scoring methods if not better results for certain dataset sizes $n$. Considering a pairwise win-rate (described in~\Cref{sec:iterative_syn_gen_results}) we can see from the row for incorrect prioritization that it is more data efficient in many instances with a high number of ``wins'' versus other methods. However at the same time looking at the corresponding column it is outperformed by many of the other methods in particular high loss and low reward selection due to its poor performance on the \GSM{} dataset so it results in a poor overall score in the final row~(\Cref{fig:win_rates_incorrect}). Overall it is a simple method and has the possibility of obtaining strong capabilities and being more data efficient than random sampling in certain domains.

\boldblue{EvoKD underperforms random sampling and other active selection methods.} Similar to Lion, which samples evenly from easy and hard data pools as determined by \llmasajudge{} scores, we can sample data evenly from correct and incorrect student predictions for synthetic data generation~\citep{liu2024evolving}. Evolving Knowledge Distillation (EvoKD), denoted as ``EvoKD'' in~\Cref{fig:learning_curves_incorrect}, can be viewed as a diversity-based sampling approach for synthetic data generation. It achieves performance comparable to incorrect-data prioritization on \GSM{} and \gotwentyfour{}, but underperforms it on \Math{} and \Pronto{}. For \GSM{}, EvoKD shares the same pathologies as promoting incorrect samples, they both underperform random sampling. EvoKD also underperforms methods that explicitly promote difficult samples~(\Cref{fig:win_rates_incorrect}), since it promotes hard samples through incorrect prioritization while simultaneously including easy samples to preserve the original data distribution. Overall, diversity-based criteria underperform approaches that emphasize difficult samples across the methods and datasets we study.

\subsection{Comparing to Other SFT Methods}
\label{sec:performance_prior_work}

\boldblue{Iterative synthetic data generation obtains comparable results to state-of-the-art SFT methods on certain datasets.} \Cref{tab:prior_works_comparison} compares the results of iterative synthetic data generation with high-loss selection to prior works in SFT which use the same LLM and similar dataset sizes. In our definition of data efficiency (\Cref{sec:prelinaries}), we can only properly compare against baselines that use the same model and that perform SFT on datasets of the same size, or if a baseline has a lower performance on a larger dataset size. Then we can conclude whether our method or the baseline is more data efficient, as defined in~\Cref{sec:prelinaries}. If a baseline has better performance with a larger dataset size, then it is not possible to say whether the baseline we are comparing against or our method is more data efficient without scaling to the same dataset sizes. Since we cannot disentangle the performance improvements due to data quality or to increased dataset sizes. For \GSM{} our work is more data efficient when compared to Orca-Math~\cite{mitra2024orca}. Also for \gotwentyfour{} our method outperforms state-of-the-art SFT baselines that use a \Qwentwopointfive{} LLM~\cite{ni2025offlinelearningforgettingreasoning}. For the \Math{} and \Pronto{} datasets we did not find a comparable SFT methods to compare data efficiency with.

\subsection{Synthetic Data Generation Preserves Properties of the Selected Data}
\label{sec:syn_data_properties}

\textboldblue{At the dataset level synthetic data generation preserves properties of the original seed dataset}. We score the selected seed dataset and take a median over scores and compare to the median score over the resulting synthetic data. If we do this for all iterations, we observe a very high rank correlation between median scores in~\Cref{fig:synthetic_data_corr}. \textboldblue{This indicates that the scores across the iterative synthetic data generation curriculum are similar before and after synthetic data generation}. 

\begin{comment}
\begin{figure}
    \centering
    \includegraphics[draft=false, width=1.0\linewidth]{plots/original_vs_synthetic_dataset_scores_median_correlation.pdf}
    \caption{\textcolor{orange}{\textbf{The median score correlation between the selected datasets and the resulting synthetic datasets}. We use a high loss (top row) and low reward (bottom row) selection algorithm and measure the median loss or reward over the datasets. We have multiple points due to considering multiple iterations of synthetic data generation and multiple seeds. The Spearman correlations ($\rho$) indicates the rank similarity of the median data scores before and after synthetic data generation over multiple iterations. The very high correlation indicates that at a dataset level synthetic data generation preserves the dataset scores. The red line shows the line of best fit to the data. The number of asterisks denote the significance level of the p-value of $\rho$: *** indicates $p < 0.001$.}}
    \label{fig:dataset_corr}
\end{figure}
\end{comment}

\begin{figure}
    \centering
    \includegraphics[draft=false, width=1.0\linewidth]{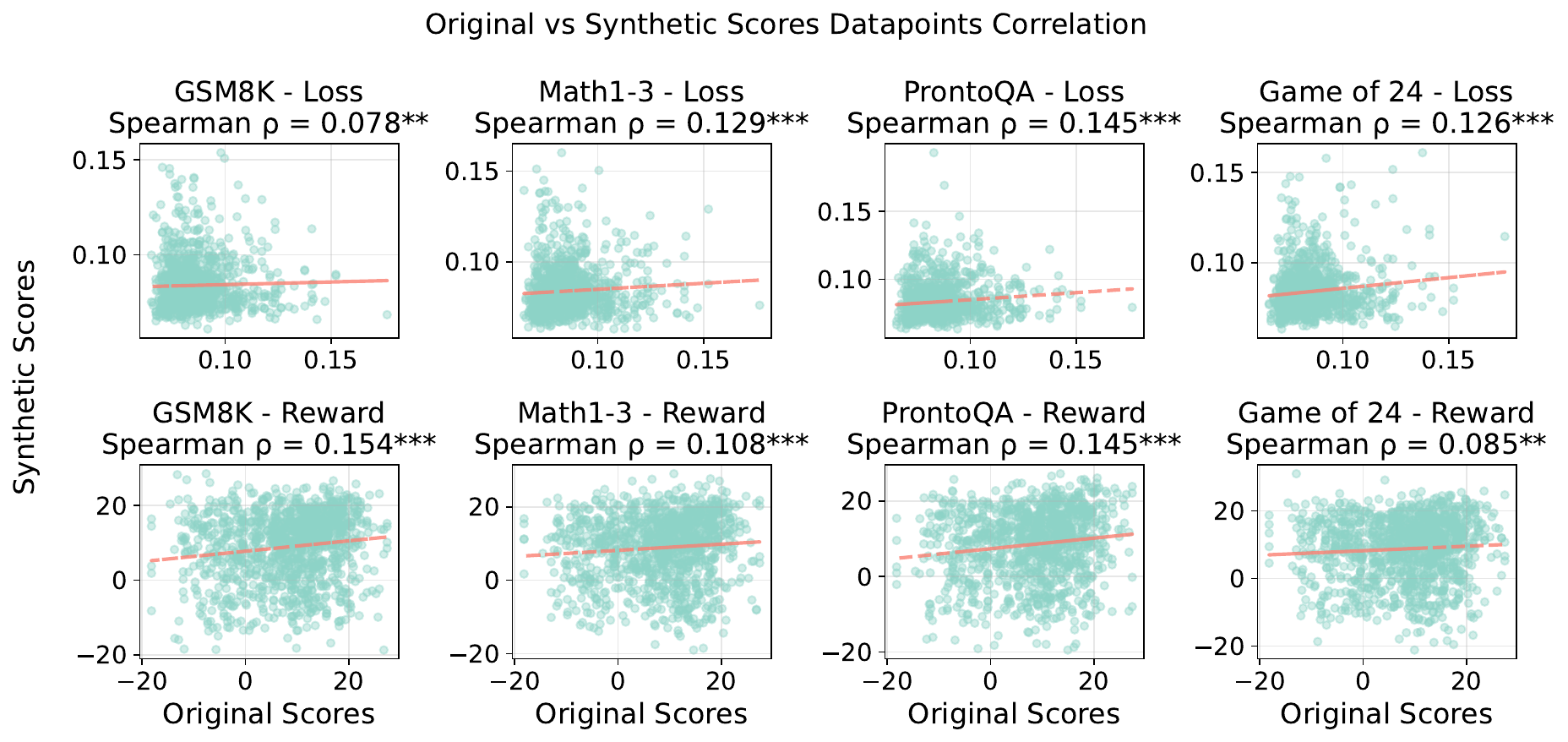}
    \caption{\textbf{The scores for individual datapoints before and after 1 step of synthetic data generation.} We consider the loss and reward of the student's predictions and look at the individual data points scores across all datasets. The Spearman correlation measures the rank correlation before and after synthetic data generation. The red line shows the line of best fit to these data. The number of asterisks denotes the rank correlation's p-value: *** indicates $p < 0.001$.}
    \label{fig:datapoint_corr}
\end{figure}

When we look at the scores over \emph{individual} data points and consider the score of a selected data point and the corresponding score of the synthetically generated datapoint, then we find there is a low but significantly greater than $0$ rank correlation between reward and loss scores for all datasets~(\Cref{fig:datapoint_corr}).

\begin{figure}
    \centering
    \includegraphics[draft=false, width=1.0\linewidth]{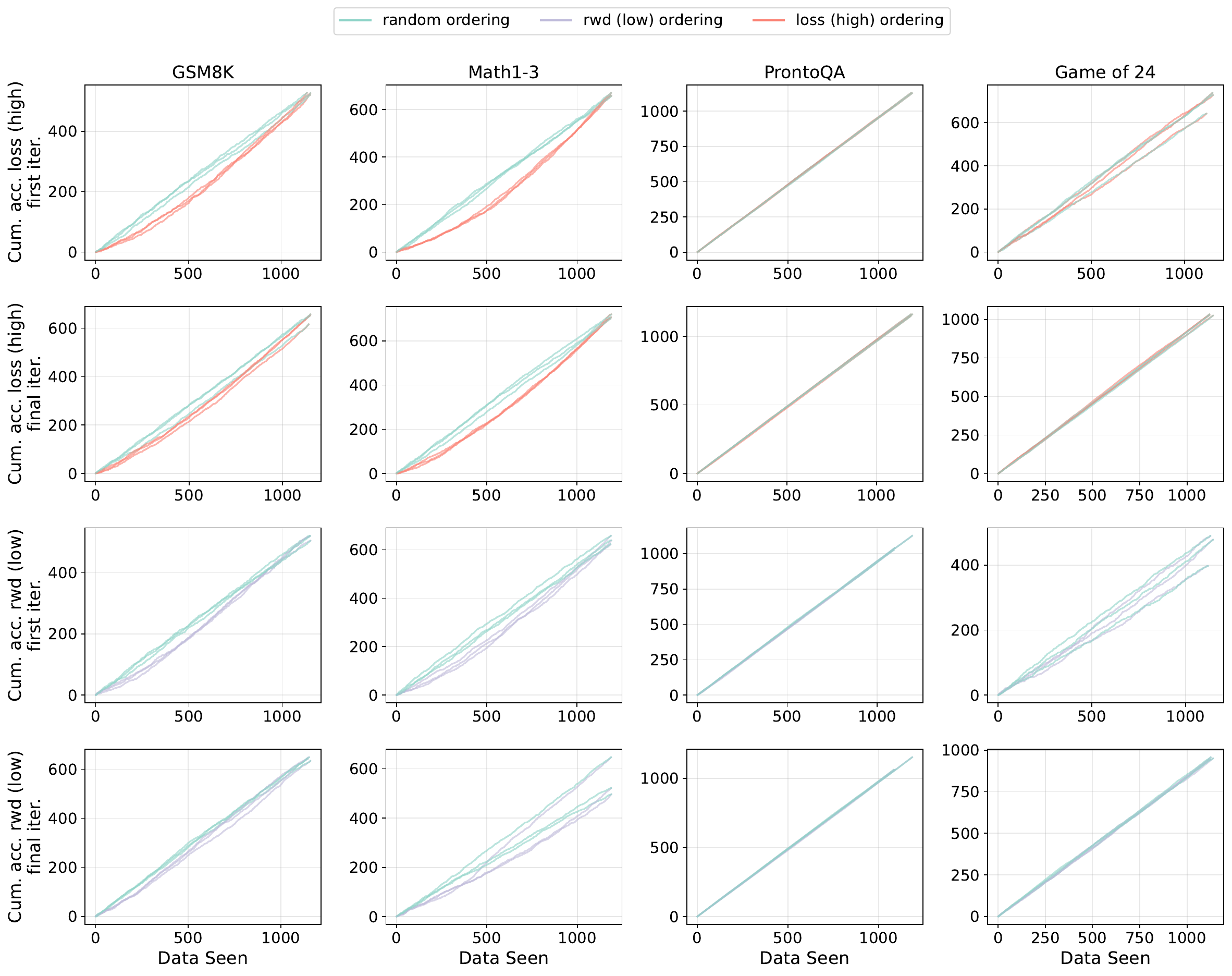}
    \caption{\textbf{The synthetic data cumulative accuracies when using random sampling and score ordering: high to low loss and low to high reward}. For each original data point we score it using the student model from the first and final iteration of iterative synthetic data generation (alternating rows). Then we generate a synthetic data point. We compare the cumulative accuracy over the synthetic data when ordering data randomly versus ordering according to the loss and reward scores. We plot individual replicates as individual lines.}
    \label{fig:cum_acc_individual_seeds}
\end{figure}

These two observations are consistent: synthetic data generation is preserving distributional factors such as dataset uncertainty (as measured by the loss over student predictions) and dataset quality (as measured by the reward over student predictions). But the noise from prompt-based synthetic data generation means that there is a low but significant correlation between scores at an individual data point level.

\subsection{Prioritizing Difficult Data Creates Difficult Synthetic Data}
\label{sec:syn_acc_cumsum}

\boldblue{The teacher produces difficult synthetic data when hard samples are prioritized by the student.} We score seed data according to its loss or reward and then generate corresponding synthetic data. We obtain the cumulative accuracy of the synthetic data ordered by the original data scores. A random ordering corresponds to random sampling, while ordering the cumulative accuracy according to a high to low loss or low to high reward corresponds to prioritizing ``difficult'' data as we do in iterative synthetic data generation. For random sampling the cumulative accuracy versus the amount of data seen so far follows a diagonal line~(\Cref{fig:cum_acc_individual_seeds}).

We plot the cumulative accuracy curves for synthetic data ordered from high to low original data loss (loss (high) ordering) in the first two rows and by low to high original data reward (rwd (low) ordering) in the final two rows of~\Cref{fig:cum_acc_individual_seeds}. For \GSM{} and \Math{} the cumulative accuracy curves for synthetic data ordered using high to low original data loss and low to high reward are below random sampling so prioritizing data according to these scores results synthetic data that the student gets lower accuracies versus random sampling. The synthetic data is ``harder'' using these active learning approaches and these ``hardness'' qualities are integrated in the synthetic data the teacher generates. This is also seen for the first iteration for the \gotwentyfour{} dataset for both scorers. In contrast, in the final iteration the student is able to get a high accuracy on the synthetic data and so it is difficult to see any difference between random ordering and prioritizing according to a high loss or low reward. This is also the case for the \Pronto{} dataset, for the first iteration we see high student accuracies for the synthetic data making comparison versus random sampling difficult, despite the reward scorer obtaining better performance than random on the \Pronto{} dataset~(\Cref{fig:learning_curves_incorrect}).

To obtain the cumulative difference plots presented in the main body of this manuscript in~(\Cref{fig:cum_acc_difference_aggregated}), we simply take the vertical distances between corresponding random sampling cumulative accuracies and the scorer cumulative accuracies in~\Cref{fig:cum_acc_individual_seeds} and aggregate across all replicates to obtain means and standard errors.

\begin{figure}
    \centering
    \includegraphics[draft=false, width=1.0\linewidth]{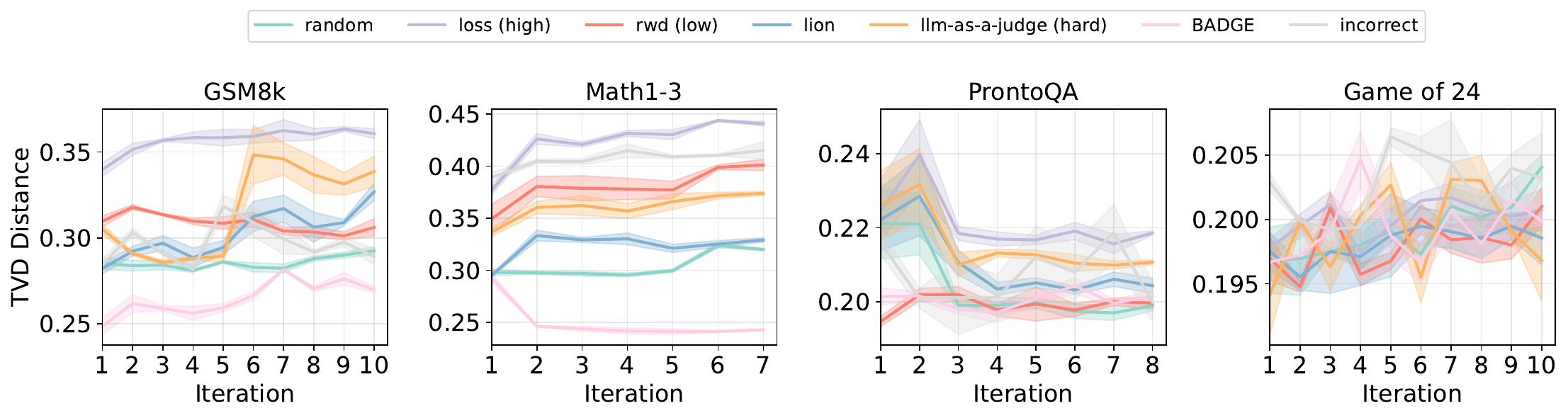}
    \caption{\textbf{The total variation distances between token distributions of our synthetic data and the original seed datasets.} We observe differences in the token distributions over the course of iterative synthetic data generation across for data selection algorithms, indicating differences in the synthetic datasets arise due to the different selection algorithms used.}
    \label{fig:tvd}
\end{figure}

\subsection{Different Selection Algorithms have their own Selection Biases}
\label{sec:tvd}

\textboldblue{The different selection algorithms we consider manifest as differences in the synthetic dataset distributions.} When we compare the synthetic datasets to the original seed datasets over the course iterative synthetic data generation, then differences between selection algorithms are evident by looking at the token distributions in~\Cref{fig:tvd}. In particular, we measure the difference between two token distributions using the total variation distance (TVD): $\text{TVD}(P_{D_0}, P_{\hat{D}_t}) = \frac{1}{2} \sum_{x\in V} |P_{D_0}(x) - P_{\hat{D}_t}(x)|$ where $x$ is a token in the vocabulary $V$ and $P$ is the empirical token distribution. The token distribution $P$ can be thought of as a histogram where the bin size is the normalized frequency of the token in the dataset. This distance is essentially looking at the absolute differences in token counts between two datasets. When measuring the TVD between synthetic datasets and the original seed dataset prior to selection, $D_0$. We can see that the distance varies between different selection algorithms which shows that there are differences in the synthetic datasets at a token distribution level. The \gotwentyfour{} dataset is the sole case where the selection algorithms yield almost indistinguishable TVDs, as its questions and answers draw from a highly restricted token range to compute 24 from four numbers using basic arithmetic operations. This points to there being distributional differences between synthetic datasets of different selection algorithms and thus shows that the selection algorithms manifest in different synthetic datasets with different properties over the course of iterative synthetic data generation. These distributional differences lead to performance differences between different selection algorithms which have been studied in the main results~(\Cref{fig:win_rates}).

\subsection{On the Design Choices for Iterative Synthetic Data Generation}
\label{sec:ablations}
\begin{figure}
    \vspace{-0.2cm}
    \centering
    \includegraphics[draft=false, width=1.0\linewidth]{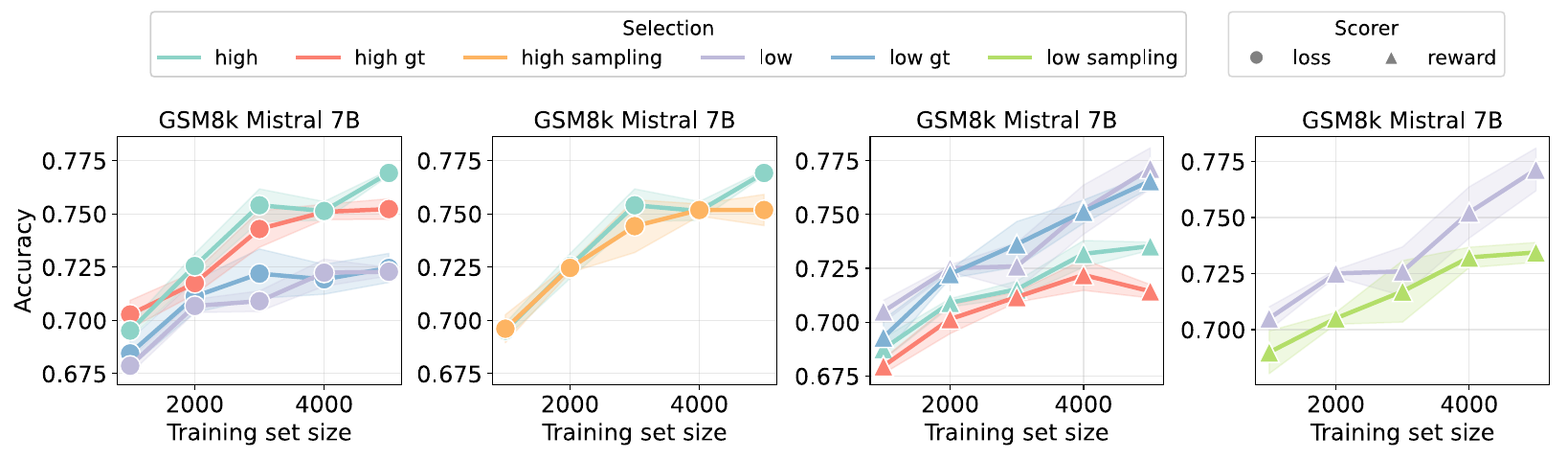}
    \caption{\textbf{Performance of iterative synthetic data generation on various data scoring and selection options}. We train on $1$k data points at each iteration with a \Mistral{} student on \GSM{}. We compare prioritizing ``difficult'' or ``easy'' data points with a high or low loss or reward. We compare using ground truth answers $y$ to the student's own predictions $\hat{y}$ and using argmax selection against sampling.}
    \label{fig:rwd_ppl_gt_sampling_ablation}
    \vspace{-0.0cm}
\end{figure}

\boldblue{Argmax selection, rather than sampling, results in the best SFT performance.} 
In~\Cref{fig:rwd_ppl_gt_sampling_ablation}, we compare various data prioritization design choices. The performance for scorers that prioritize data where the student answer is the most uncertain (high loss) or worse quality (low reward) results in the best performance when compared to data for which the model is confident (low loss) or is of better quality (high reward). Furthermore, we compare whether using the ground truth answer $y$ (denoted ``gt'' in~\Cref{fig:rwd_ppl_gt_sampling_ablation}) or the student's own prediction $\hat{y}$ is more data efficient. We can see worse performance when computing scores with the ground-truth answer for the loss scorer, while scoring with the reward model results in equal SFT performance. \boldblue{There is no benefit to using the ground truth answers over the student's own predictions.}

\begin{wrapfigure}{r}{0.5\linewidth}
    \centering
    \vspace{-1.5cm}
    \includegraphics[draft=false, width=\linewidth]{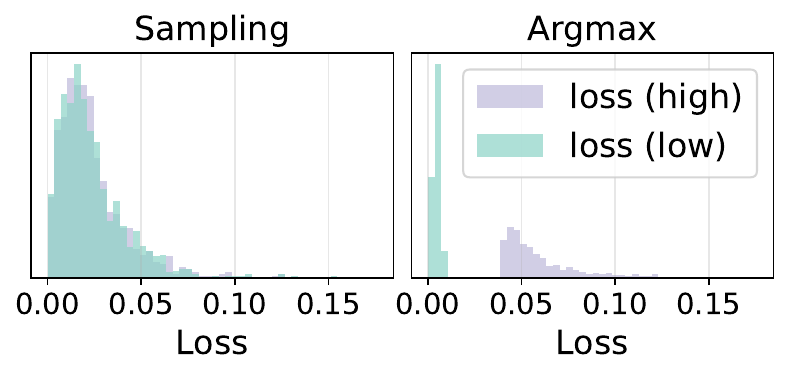}
    \vspace{-0.8cm}
    \caption{\textbf{Distribution of losses for different sampling methods}. We select $1$k according to a high or low loss sampling (left) and argmax selection (right) for \GSM{} and can see almost no difference when using sampling.}
    \vspace{-1.0cm}
    \label{fig:sampling_vs_greedy}
\end{wrapfigure}

Finally, we compare selection methods: argmax selection and sampling and can see lower SFT performance when using sampling (labelled with ``sampling'' in~\Cref{fig:rwd_ppl_gt_sampling_ablation}). We sample $m$ points by sampling from a distribution proportional to these scores: $\bar{D}_t \overset{m}{\sim}{\text{softmax}}(\left \{s_i\right\}_{i=1}^{n})$. We found poor performance when sampling because sampling from the softmax distribution of loss or reward scores results in a similar distribution of scores for selected data as if we performed random sampling. Moreover, if we select $m=1$k data points from the \GSM{} seed dataset and look at the distribution of loss scores via sampling for the highest and lowest $1$k scoring data, then the distributions are indistinguishable to the naked eye. Argmax selection however produces distinct distributions~(\Cref{fig:sampling_vs_greedy}).

\section{Dataset Further Details}
\label{sec:dataset_further_details}

In this section we provide in depth details on the datasets used in our experiments together with the dataset sizes used throughout our empirical study of iterative synthetic data generation~(\Cref{sec:seed_datasets}). Also we provide the teacher prompts used for synthetic data generation~(\Cref{sec:prompts}).

We introduce the seed question and answer datasets $D_0$. The validation and test sets are taken from the original seed datasets as opposed to using synthetic data. The train sets $\hat{D}_t$ are synthetically generated. We summarize the datasets sizes in~\Cref{sec:seed_datasets}. Unless otherwise stated we use a \GPTfouro{} teacher. We prompt the teacher with few-shot examples from $D_0$ to generate a new synthetic questions~\citep{liu2024evolving}. For all datasets we throw away similar synthetic questions if the rouge-score~\citep{lin-2004-rouge} with respect to all previously generated questions is above $0.7$~\citep{jiang2023lion}.

\paragraph{\GSM.} We perform SFT on a \Mistral{}~\citep{jiang2023mistral} student on school level mathematics questions~\citep{cobbe2021training}. We use an external language model $\texttt{gpt4o-mini}$ to assess whether the student's answer is equivalent to the ground truth answer, in a similar manner to~\citet{mitra2024orca}, see~\Cref{sec:eval_prompts} for prompting details. We take $748$ question-answer pairs from the test set as a validation set and use $500$ question-answer pairs as a test set\footnote{\url{https://huggingface.co/datasets/openai/gsm8k}}.

\paragraph{\Math.} We finetune a \Llama{}~\citep{dubey2024llama} student on the competition math dataset~\citep{hendrycksmath2021} which consists of more difficult math questions\footnote{\url{https://huggingface.co/datasets/hendrycks/competition_math}}. The dataset is classified into $5$ levels of question difficulty. We use the easiest levels $1$ to $3$ and pick $500$ question-answer pairs from the test set for validation. We assess the correctness of an answer by matching the solution to the regular expression $\texttt{\textbackslash boxed\{(\textbackslash d*)\}}$. The dataset is also categorized by the type of mathematics question: geometry, algebra etc. We use the category in our synthetic data generation prompt.

\paragraph{\Pronto.} The questions are synthetically generated logical chain-of-thought style reasoning questions with boolean answers~\citep{saparov2022language}. We perform SFT with a \Qwenonepointfive{} student model. We use an external language model $\texttt{gpt4o-mini}$ to assess whether the student's reasoning steps are correct and whether the student answer is equivalent to the ground truth answer like for \GSM{}, see~\Cref{sec:eval_prompts} for details. We use $300$ question-answer pairs as a validation set and the remaining $200$ as a test set\footnote{We use\url{https://huggingface.co/datasets/renma/ProntoQA} for validation and testing, as a train set we use \url{https://huggingface.co/datasets/longface/prontoqa-train} like in~\citep{huang2024self}, questions are distinct between these two ProntoQA datasets.}.

\paragraph{\gotwentyfour.} We use a \Qwentwopointfive{}~\citep{qwen2025qwen25technicalreport} student for SFT on the task of using $4$ numbers to obtain the number $24$ by finding which basic arithmetic operations are needed\footnote{\url{https://huggingface.co/datasets/nlile/24-game}}. Each question can have multiple solutions, we treat each solution as a separate data point. We use backward reasoning to synthetically generate new questions~\citep{jiang2023forward} and use \GPTothree{} as a teacher model (qualitatively this produces better questions than \GPTfouro{}). In backward reasoning if the answer is $\texttt{13*8-10*8=24}$, for example, we can construct a new question by setting two integers to variables $\texttt{a*b-10*8=24}$ and solving to generate new questions and answers~\citep{jiang2023forward}. We verify that the backward reasoned final answer evaluates to $24$ and that it uses the $4$ numbers in the question. We use \GPTfouro{} to then generate reasoning steps to obtain the final backward-reasoned answer. We assess the correctness of the student's final answer by matching the regular expression in $\texttt{\textbackslash boxed\{\}}$ and that the extracted answer evaluates to $24$ and checking that all numbers in the question are used once. Synthetic questions are not checked for rouge-score overlap since the set of tokens required to make questions and answers is a small subset of the vocabulary.

\begin{table}[ht]
  \centering
  \begin{tabular}{l c c c}
    \toprule
    \textbf{Dataset} & \textbf{Seed Size} & \textbf{Validation Size} & \textbf{Test size}\\
    \midrule
    \GSM{} & 7473 & 748 & 500 \\
    \Math{} & 3504 &  500 & 500\\
    \Pronto & 2880 & 300 & 200\\
    \gotwentyfour & 2217 & 500 & 300 \\
    \bottomrule
  \end{tabular}
  \caption{Summary of the seed dataset sizes, validation and test set sizes. For all datasets we use $1$k data points per iteration for finetuning.}
  \label{tab:seed_dataset_sizes}
\end{table}

\subsection{Seed Dataset Sizes}
\label{sec:seed_datasets}

We summarize the seed dataset sizes for all datasets used in our experiments in~\Cref{tab:seed_dataset_sizes}. The seed dataset $D_0$, is used for scoring and selecting data points to get the selected data $\bar{D}_t$. The selected data is then put forward for prompt-based synthetic data generation~(\Cref{sec:prompt_syn_gen}). We set the validation and test sets to be from the original seed datasets. We use the resulting synthetic datasets $\hat{D}_t$ for SFT, we generate a fixed sized synthetic dataset to enable fair comparison between selection methods and assess data efficiency~(\Cref{sec:datasets}).

\subsection{Synthetic Data Generation Prompts}
\label{sec:prompts}

We provide the prompts used for prompt-based synthetic data generation (described in~\Cref{sec:prompt_syn_gen}) below for all datasets used in our experiments:
\begin{itemize}
    \item \GSM{} see~\Cref{sec:prompt_gsm}.
    \item \Math{} see~\Cref{sec:prompt_math}.
    \item \Pronto{} see~\Cref{sec:prompt_pqa}.
    \item \gotwentyfour{} see~\Cref{sec:prompt_go24}.
\end{itemize}

\subsubsection{Grade School Maths}
\label{sec:prompt_gsm}

Below is the prompt we use for synthetic question generation for \GSM{} using a~\GPTfouro{} teacher. In the prompt below $\{0\}$ are few-shot examples of questions and answers: $\{z_i\}_{i=1}^k \sim D_0$, we set $k=5$ for all our experiments and $\{1\}$ is the question from the data selected by the student: $\bar{x} = \bar{z}[0]$ where $\bar{z} \sim \bar{D}_t$. The few-shot examples are formatted as follows: $\texttt{\#Given Instruction\#: \{\} \#Answer\#: \{\}}$

\begin{tcolorbox}[colframe=black!50, colback=gray!10, title=\GSM{} synthetic question generation prompt]
\vspace{0cm} % Adjust space inside the box
\texttt{I want you to act as Instruction Creator. \\
Your objective is to rewrite a \#Given Instruction\# into a more complex version, to make it a bit harder. \\
The \#Rewritten Instruction\# must be reasonable and must be understood and responded to by humans.\\
Here are some \#Examples\#:\\
$\{0\}$\\
I want you to act as Instruction Creator. \\
Your objective is to rewrite a \#Given Instruction\# into a more complex version, to make it a bit harder. \\
The \#Rewritten Instruction\# must be reasonable and must be understood and responded to by humans.\\
You MUST complicate the \#Given Instruction\# using the following method: \\
1. Change the names of people \#Given Instruction\#.\\
2. Change the objects in the \#Given Instruction\#. \\
3. Change any quantities and durations in the \#Given Instruction\#.\\
4. Add 1 to 3 more operations in \#Rewritten Instruction\#.\\
5. Change the operations, for example: multiplication, division, subtraction, addition, percentages, fractions and combinations of these. \\
6. You should try your best not to make the \#Rewritten Instruction\# become verbose, \#Rewritten Instruction\# can only add 10 to 20 words into \#Given Instruction\#.\\
Use \#Examples\# to complicate \#Given Instruction\#.\\
'\#Given Instruction\#', '\#Rewritten Instruction\#', 'given instruction' and 'rewritten instruction' are not allowed to appear in \#Rewritten Instruction\#.\\
\#Given Instruction\#: \\
$\{1\}$\\
\#Rewritten Instruction\#:}
\vspace{0cm}
\end{tcolorbox}

We use the following prompt to obtain synthetic answers from our \GPTfouro{} teacher (and from our student model):
\begin{tcolorbox}[colframe=black!50, colback=gray!10, title=\GSM{} answer prompt]
\vspace{0cm} % Adjust space inside the box
\texttt{Question: \{\} Solve the problem step-by-step. Answer:}
\vspace{0cm}
\end{tcolorbox}

\subsubsection{Math1-3}
\label{sec:prompt_math}
Below is the prompt we use for synthetic question generation for \Math{} using a~\GPTfouro{} teacher, $\{0\}$ are few shot examples of questions, answers and the type of problem e.g. Geometry, Algebra etc. The number of few-shot examples is $5$ and are of the same type as the seed question. In the prompt below $\{1\}$ is the type of mathematics problem and $\{2\}$ is the question from the selected dataset: $\bar{x} = \bar{z}[0]$ where $\bar{z} \sim \bar{D}_t$. The few-shot examples are formatted as follows: $\texttt{The type of math problem is \{\}. \#Given Instruction\#: \{\}}$ $\texttt{ \#Answer\#: \{\}}$

\begin{tcolorbox}[colframe=black!50, colback=gray!10, title=\Math{} synthetic question generation prompt]
\vspace{0cm} % Adjust space inside the box

\texttt{I want you to act as an Instruction Creator for \{1\} mathematics problems. \\
Create a new question \#Rewritten Instruction\# by using \#Given Instruction\# as inspiration. The new question should have a single unique answer. \\
Ensure that the type of the question you generate \#Rewritten Instruction\# matches the type of instruction \#Given Instruction\#. \\
Make \#Rewritten Instruction\# different from \#Given Instruction\#. \\
The \#Rewritten Instruction\# must be reasonable, have a solution and must be understood and responded to by humans.
Here are some \#Examples\#: \\
\{0\} \\
Use \#Examples\# as inspiration to make \#Rewritten Instruction\# different to \#Given Instruction\#. \\
'\#Given Instruction\#', '\#Rewritten Instruction\#', 'given instruction' and 'rewritten instruction' are not allowed to appear in \#Rewritten Instruction\#. \\
\#Given Instruction\# is a \{1\} math problem. \\
\#Given Instruction\#: \\
\{2\} \\
\#Rewritten Instruction\#: \\
}
\vspace{0cm}
\end{tcolorbox}

We use the following prompt for obtaining synthetic answers from our \GPTfouro{} teacher (and for obtaining answers from our student model):
% \begin{verbatim}
% "Can you solve the following math problem? \{0\}. Provide a bullet point summary 
% of your reasoning. Your final answer should be a single answer, in the form 
% \textbackslash boxed\{answer\}, at the end of your response."
% \end{verbatim}
\begin{tcolorbox}[colframe=black!50, colback=gray!10, title=Math1-3 answer prompt]
\vspace{0cm} % Adjust space inside the box
\texttt{Can you solve the following math problem? \{0\}. Provide a bullet point summary\
of your reasoning. Your final answer should be a single answer, in the form\
\textbackslash boxed\{answer\}, at the end of your response.}
\vspace{0cm}
\end{tcolorbox}

\subsubsection{ProntoQA}
\label{sec:prompt_pqa}
We present the prompt we use for synthetic question generation using a~\GPTfouro{} teacher for the \Pronto{} dataset~\citep{saparov2022language}. A datapoint from the \Pronto{} dataset is comprised of a context, question and answer $z = (x = (c, q), y)$ where $x$ is comprised of the context $c$ and question $q$. The answers $y$ are boolean. The few-shot question generation is therefore comprised of contexts and questions for the teacher to generate new synthetic context and questions, $\hat{x}$. In the prompt below $\{0\}$ are few-shot examples of questions and answers from $\{z_i\}_{i=1}^k \sim D_0$, we set $k=5$ for all our experiments and $\{1\}$ is the question from the selected dataset $\bar{x} = \bar{z}[0]$ where $\bar{z} \sim \bar{D}_t$. The few-shot examples $\{0\}$ are formatted as follows: $\texttt{Context: \{\} Question: \{\}}$.

\begin{tcolorbox}[colframe=black!50, colback=gray!10, title=ProntoQA synthetic question generation prompt]
\vspace{0cm} % Adjust space inside the box

\texttt{I want you to act as an Instruction Creator for logical problems.\\
Create a new question \#Rewritten Instruction\# by using \#Given Instruction\# as inspiration.\\
Make \#Rewritten Instruction\# different from \#Given Instruction\# by changing the names, objects and adjectives. 
Also vary the number of logical reasoning steps in \#Rewritten Instruction\#. Ensure that it is possible to answer the question with true or false answer.\\
The \#Rewritten Instruction\# must be reasonable, have a solution and must be understood and responded to by humans.\\
Here are some \#Examples\#:\\
\{0\}\\
Use \#Examples\# as inspiration to make \#Rewritten Instruction\# different to \#Given Instruction\#.\\
'\#Given Instruction\#', '\#Rewritten Instruction\#', 'given instruction' and 'rewritten instruction' are not allowed to appear in \#Rewritten Instruction\#.\\
\#Given Instruction\#: \\
\{1\}\\
\#Rewritten Instruction\#:\\}
\vspace{0cm}
\end{tcolorbox}

We use the following prompt for obtaining synthetic answers from the \GPTfouro{} teacher (and for obtaining answers from our student model):

\begin{tcolorbox}[colframe=black!50, colback=gray!10, title=\Pronto{} answer prompt]
\vspace{0cm} % Adjust space inside the box
\texttt{Context: \{\} Let's think step by step. Response:}
\vspace{0cm}
\end{tcolorbox}

\subsubsection{Game of 24}
\label{sec:prompt_go24}
Below is the prompt we use for synthetic question generation using~\GPTothree{} teacher for the \gotwentyfour{} dataset. A datapoint from the \gotwentyfour{} dataset is comprised of a set of four numbers and the arithmetic one-line solution to obtain $24$. In the prompt below $\{0\}$ is the question, a set of numbers for instance $\bar{x} = [8,8,10,12]$ and $\{1\}$ is the arithmetic answer for instance $\bar{y} = (12-10)\times8 + 8$ where  $\bar{z} = (\bar{x}, \bar{y})$ and $\bar{z} \sim \bar{D}_t$. We use backward reasoning to to obtain a new question and answer to the \gotwentyfour{} (see the prompt below). We verify that the synthetic answer evaluates to $24$ and that all the numbers from the synthetic question are also present in the synthetic answer. Since backward reasoning for synthetic data generation produces both the question and the answer, we then prompt our teacher, \GPTfouro{} in a second step, with both the synthetic question and answer to get a synthetic reasoning trace without any verification of the reasoning steps to construct our synthetic dataset $\hat{D}_{t}$ (in the second prompt below).

\begin{tcolorbox}[colframe=black!50, colback=gray!10, title=Game of 24 synthetic question generation prompt]
\vspace{0cm} % Adjust space inside the box

\texttt{I want you to act as an instruction creator. I want you to write a new problem to the game of 24. \\
The numbers \{0\} need to be used to obtain the number 24. Use each number once, even if a number is repeated use it multiple times, with the arithmetic operations +, -, *, / to obtain 24.
Here is how the above numbers \{0\} are used to obtain 24: \{1\}. \\
\\
I want you to create a new problem to the game of 24 using \{1\}. Let's use a backward thinking method. Take two of the distinct numbers in \{1\}. Call them a and b. Then construct an equation with two unknowns, a and b. Pick integer values for the first variable b then solve for a. \\
\\
For example the numbers 8, 8, 10, 13 can be used to get 24: 13*8-10*8=24. We can construct the following equation a*b-10*8=24 by substituting a=13 and b=8. Rearranging we get a=104/b. Let's pick an integer which divides into 104 for b: b=4 therefore a=26.\\
We also could have picked b=2 and so a=62. Therefore one possible answer to the game of 24 using this backward method is \textbackslash boxed\{4*26-10*8\}. If no answer is possible return \textbackslash boxed\{null\}. \\
\\
Here is the current solution \{1\} again. Enclose the new equation which results in 24 in \textbackslash boxed\{\}. Let's use this backward thinking method and think step by step. \\}

\vspace{0cm}
\end{tcolorbox}

\begin{tcolorbox}[colframe=black!50, colback=gray!10, title=Game of 24 prompt for synthetic reasoning steps]
\vspace{0cm} % Adjust space inside the box

\texttt{Use numbers and basic arithmetic operations (+ - * /) to obtain 24. Each step, you are only allowed to choose two of the remaining numbers to obtain a new number.\\
Input: 4 4 6 8\\
Steps: \\
4 + 8 = 12 (left: 4 6 12)\\
6 - 4 = 2 (left: 2 12)\\
2 * 12 = 24 (left: 24)\\
Answer: (6 - 4) * (4 + 8) = 24\\
Input: 2 9 10 12\\
Steps:\\
12 * 2 = 24 (left: 9 10 24)\\
10 - 9 = 1 (left: 1 24)\\
24 * 1 = 24 (left: 24)\\
Answer: (12 * 2) * (10 - 9) = 24
Input: 4 9 10 13\\
Steps:\\
13 - 10 = 3 (left: 3 4 9)\\
9 - 3 = 6 (left: 4 6)\\
4 * 6 = 24 (left: 24)\\
Answer: 4 * (9 - (13 - 10)) = 24\\
Input: 1 4 8 8\\
Steps:\\
8 / 4 = 2 (left: 1 2 8)\\
1 + 2 = 3 (left: 3 8)\\
3 * 8 = 24 (left: 24)\\
Answer: (1 + 8 / 4) * 8 = 24\\
Input: 5 5 5 9\\
Steps:\\
5 + 5 = 10 (left: 5 9 10)\\
10 + 5 = 15 (left: 9 15)\\
15 + 9 = 24 (left: 24)\\
Answer: ((5 + 5) + 5) + 9 = 24\\
Input: \{question\}\\
Here is the final answer: \{answer\}\\
Provide the steps to obtain the final answer which equates to 24, as if you did not have access to the answer. Put your final answer within \textbackslash boxed\{answer\}. Steps:
}

\vspace{0cm}
\end{tcolorbox}

We use the following prompt to get answers from the student, similarly to~\citet{ni2025offlinelearningforgettingreasoning}:

\begin{tcolorbox}[colframe=black!50, colback=gray!10, title=Game of 24 student prediction prompt]
\vspace{0cm} % Adjust space inside the box

\texttt{Use numbers and basic arithmetic operations (+ - * /) to obtain 24. Each step, you are only allowed to choose two of the remaining numbers to obtain a new number.\\
Input: 4 4 6 8\\
Steps: \\
4 + 8 = 12 (left: 4 6 12)\\
6 - 4 = 2 (left: 2 12)\\
2 * 12 = 24 (left: 24)\\
Answer: (6 - 4) * (4 + 8) = 24\\
Input: 2 9 10 12\\
Steps:\\
12 * 2 = 24 (left: 9 10 24)\\
10 - 9 = 1 (left: 1 24)\\
24 * 1 = 24 (left: 24)\\
Answer: (12 * 2) * (10 - 9) = 24
Input: 4 9 10 13\\
Steps:\\
13 - 10 = 3 (left: 3 4 9)\\
9 - 3 = 6 (left: 4 6)\\
4 * 6 = 24 (left: 24)\\
Answer: 4 * (9 - (13 - 10)) = 24\\
Input: 1 4 8 8\\
Steps:\\
8 / 4 = 2 (left: 1 2 8)\\
1 + 2 = 3 (left: 3 8)\\
3 * 8 = 24 (left: 24)\\
Answer: (1 + 8 / 4) * 8 = 24\\
Input: 5 5 5 9\\
Steps:\\
5 + 5 = 10 (left: 5 9 10)\\
10 + 5 = 15 (left: 9 15)\\
15 + 9 = 24 (left: 24)\\
Answer: ((5 + 5) + 5) + 9 = 24\\
Input: \{question\}\\
Put your final answer within \textbackslash boxed\{answer\}. Steps:
}

\vspace{0cm}
\end{tcolorbox}

\subsection{Evaluation prompts}
\label{sec:eval_prompts}

To assess whether the student's prediction is equal to the ground-truth answer we use $\texttt{gpt4o-mini}$ to verify the correctness of the student. We use the following prompt and a system prompt which is different for each dataset used:

\begin{tcolorbox}[colframe=black!50, colback=gray!10, title=\GSM{} and \Pronto{} evaluation prompt]
\vspace{0cm} % Adjust space inside the box
\texttt{Question:\{\} Problem Setter’s answer:\{\} Student’s answer:\{\}}
\vspace{0cm}
\end{tcolorbox}

For \GSM{} we use the following system prompt for evaluation, similarly to~\citet{mitra2024orca}:

\begin{tcolorbox}[colframe=black!50, colback=gray!10, title=\GSM{} evaluation system prompt]
\vspace{0cm} % Adjust space inside the box

\texttt{As an expert Math teacher, your role is to evaluate a student’s answer to a word problem. The problem is accompanied by a correct solution provided by the problem setter. It is important to remember that there may be various methods to solve a word problem, so the student’s steps might not always align with those in the problem setter’s solution. However, the final answer, typically a number, should be unique and match the problem setter’s answer. Your task involves analyzing the student’s solution to identify any mistakes and determine whether the answer can be modified to correct the error. If the student’s answer is unfixable, consider creating practice problems to help improve their understanding. Use the following format: Error Analysis: In one sentence, extract the final answer from the problem setter’s solution and compare it with the student’s answer. Do they match? Final Verdict: Correct/Incorrect.}

\vspace{0cm}
\end{tcolorbox}

For \Pronto{} we use the following system prompt for evaluation:

\begin{tcolorbox}[colframe=black!50, colback=gray!10, title=\Pronto{} evaluation system prompt]
\vspace{0cm} % Adjust space inside the box
\texttt{You are a logical expert. Your role is to evaluate a student’s answer to a logical reasoning problem. The problem is accompanied by a correct solution provided by the problem setter. Your task is to assess whether the problem setter's answer and the student's answer match. Use the following format: Error Analysis: In one sentence, extract the final answer from the problem setter’s solution and compare it with the student’s answer. Do they match? Final Verdict: Correct/Incorrect.}
\vspace{0cm}
\end{tcolorbox}

If the output contains string variations of $\texttt{"Final Verdict: Correct"}$ then the student's prediction is correct and wrong otherwise. For the \Math{} and \gotwentyfour{} datasets we use pattern matching to extract the student's answer and compare to the ground truth, see~\Cref{sec:datasets} for details.

\end{document}